\documentclass[10pt,twocolumn,letterpaper]{article}

\usepackage{comment}
\usepackage{tabu}
\usepackage[pagenumbers]{cvpr} 

\usepackage[dvipsnames]{xcolor}

\usepackage[ruled,linesnumbered]{algorithm2e}
\usepackage{algorithmic}

\usepackage{amssymb}
\usepackage{pifont}
\usepackage{colortbl}

\usepackage{sidecap}

\usepackage{amsmath,amsfonts,bm,amsthm}
\usepackage{mathtools}

\newcommand{\R}{\mathbb{R}}
\newcommand{\E}{\mathbb{E}}

\newcommand{\dec}{\mathcal{D}}
\newcommand{\enc}{\mathcal{E}}

\newtheorem{theorem}{Theorem}[section]
\newtheorem{proposition}[theorem]{Proposition}

\newtheorem{assumption}[theorem]{Assumption}

\def\mA{{\mathbf{A}}}

\def\mI{{\mathbf{I}}}

\def\mX{{\mathbf{X}}}

\def\gA{{\mathcal{A}}}
\def\gB{{\mathcal{B}}}
\def\gC{{\mathcal{C}}}
\def\gD{{\mathcal{D}}}

\def\gG{{\mathcal{G}}}

\def\gJ{{\mathcal{J}}}

\def\gL{{\mathcal{L}}}

\def\gN{{\mathcal{N}}}
\def\gO{{\mathcal{O}}}

\def\vzero{{\mathbf{0}}}

\def\vmu{{\mathbf{\mu}}}
\def\vtheta{{\mathbf{\theta}}}
\def\veps{{\mathbf{\epsilon}}}

\def\vn{{\mathbf{n}}}

\def\vx{{\mathbf{x}}}
\def\vy{{\mathbf{y}}}
\def\vz{{\mathbf{z}}}

\definecolor{cvprblue}{rgb}{0.21,0.49,0.74}
\usepackage[pagebackref,breaklinks,colorlinks,citecolor=cvprblue]{hyperref}

\newcommand\vincent[1]{\textcolor{red}{[Vincent: #1]}}

\def\paperID{17512} 
\def\confName{CVPR}
\def\confYear{2024}

\title{Beyond First-Order Tweedie: Solving Inverse Problems using Latent Diffusion
\vspace{-1ex}
}

\author{
Litu Rout$^{1,2,*}$,
Yujia Chen$^{1}$,
Abhishek Kumar$^{3}$,\\
Constantine Caramanis$^{2}$,
Sanjay Shakkottai$^{2}$,
Wen-Sheng Chu$^{1}$\\
$^1$ Google Research, $^2$ UT Austin, $^3$ Google DeepMind\\
{\tt\small\{litu.rout,constantine,sanjay.shakkottai\}@utexas.edu
\{yujiachen,abhishk,wschu\}@google.com}
}

\begin{document}

\twocolumn[{%
\renewcommand\twocolumn[1][]{#1}%
\maketitle
\vspace{-7ex}
\begin{center}
    \includegraphics[width=\textwidth]{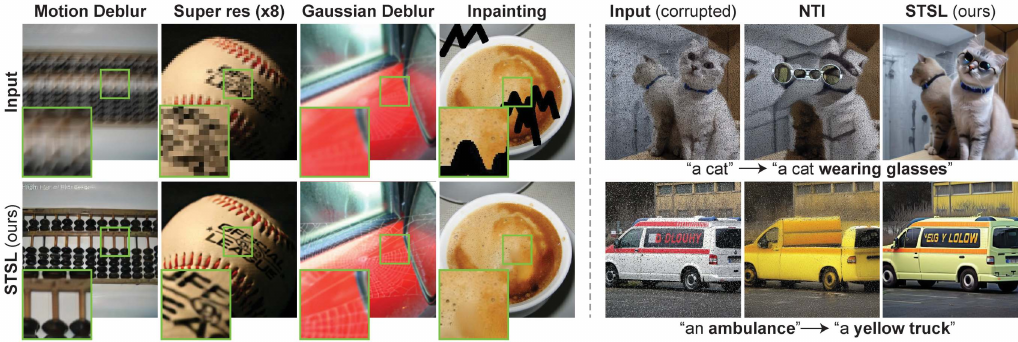}
    \vspace{-4ex}
    \captionof{figure}{
    {\bf Image inversion and editing with latent diffusion:}
    Our method, termed STSL, provides efficient inversion while enhancing quality of reconstructed images, even when addressing corruptions (\eg, blurriness, low resolution, noise).
    We show its versatility in various inversion tasks ({\bf left figure}): motion deblurring, super-resolution, gaussian deblurring, and inpainting.
    In addition, STSL extends to text-guided image editing with corrupted images ({\bf right}), surpassing the performance of NTI \cite{nti}, a prominent method in this domain. 
    }
    \label{fig:main}
    \vspace{0ex}
\end{center}%
}]

\begin{abstract}
\vspace{-3.5ex}
Sampling from the posterior distribution poses a major computational challenge in solving inverse problems using latent diffusion models.
Common methods rely on Tweedie's first-order moments, which are known to induce a quality-limiting bias~\citep{meng2021estimating}.
Existing second-order approximations are impractical due to prohibitive computational costs, making standard reverse diffusion processes intractable for posterior sampling.
This paper introduces Second-order Tweedie sampler from Surrogate Loss (STSL), a novel sampler that offers efficiency comparable to first-order Tweedie with a tractable reverse process using second-order approximation.
Our theoretical results reveal that the second-order approximation is lower bounded by our surrogate loss that only requires $\gO(1)$ compute using the trace of the Hessian, and by the lower bound we derive a new drift term to make the reverse process tractable.
Our method surpasses SoTA solvers PSLD \cite{psld} and P2L \cite{p2l}, achieving 4X and 8X reduction in neural function evaluations, respectively, while notably enhancing sampling quality on FFHQ, ImageNet, and COCO benchmarks.
In addition, we show STSL extends to text-guided image editing and addresses residual distortions present from corrupted images in leading text-guided image editing methods.
To our best knowledge, this is the first work to offer an efficient second-order approximation in solving inverse problems using latent diffusion  and editing real-world images with corruptions.
\end{abstract}

\vspace{-5ex}
\section{Introduction}
\label{sec:intro}
\vspace{0ex}

This paper focuses on solving inverse problems using pre-trained latent diffusion models. 
The goal of inverse problem solvers is to find an image $\vx\in \R^d$ that satisfies $\vy=\mA\vx+\vn,~\vn\sim \gN\left(\vzero,\sigma_\vy^2\mI_d\right)$\footnote{
    \textbf{Notations.} Bold capital letter $\mX$, bold small letter $\vx$, normal capital letter $X$, and normal small letter $x$ denote a matrix, a vector, a vector-valued random variable, and a scalar respectively. 
    The dimension of the identity matrix $\mI$ should be apparent from the context. 
    $^*$ This work was done during an internship at Google.
}
, where $\mA\in \R^{k\times d}$ is a measurement operator and $\vy\in \R^k$ is a noisy observation.
This gives rise to a sampling challenge, where the objective is to sample the posterior $p(X|Y=\vy)$.
Diffusion models are gaining popularity as priors ($p_t(X_t)$) for solving inverse problems~\citep{chung2022improving,dps,pGDM,psld,p2l}.
However, the likelihood term ($p_t(\vy|X_t)$) 
is only available for $t=0$, but not for $t>0$, making posterior sampling inconsistent with the Bayesian posterior.
One way to address this issue is training a noise conditional likelihood model, yet this is limited by training costs and the need for re-training when the measurement operator $\mA$ changes.
State-of-the-art methods, such as PSLD~\cite{psld} and P2L~\cite{p2l},
resort to alternatives for computing $p_t(\vy|X_t)$.
Among these methods, Tweedie's formula with first-order moments is commonly used to obtain a conditional expectation of the clean image ($X_0$) given the noisy image ($X_t$), \ie, $\E_{X_0 \sim p_t\left(X_0|X_t \right)}\left[ X_0\right]$.
The expected clean image is then used to approximate the likelihood as $p_t(\vy|X_t)=\E_{X_0 \sim p_t\left(X_0|X_t \right)}p_t(\vy|X_0)\approx p_t(\vy|\E_{X_0 \sim p_t\left(X_0|X_t \right)}\left[ X_0\right])$. 
This step is crucial in diffusion-based inverse problem solvers~\citep{dps,psld,p2l}.

Samplers relying on Tweedie's first-order moments are prone to sub-optimal performance due to biases in reconstruction~\citep{meng2021estimating,menon2020pulse, ajil_bias, mri_paper}.
Recent efforts have aimed to address this bias and improve the results by introducing second-order approximation using Tweedie's formula \citep{meng2021estimating,tmpd}. 
Despite these attempts, the first-order approximation is still widely used in SoTA solvers~\citep{psld,p2l}, as existing second-order alternatives~\citep{meng2021estimating,tmpd} are hindered by significant time or memory complexity and make conventional reverse diffusion processes intractable for posterior sampling.
As a result, it remains  relatively unexplored to solve inverse problems with Tweedie's second-order approximation.

In this paper, we introduce Second-order Tweedie sampler from Surrogate Loss (STSL), offering a new surrogate loss function that gives rise to a tractable reverse diffusion process using an efficient second-order approximation (\S\ref{sec:method}). 
Crucially, updating the drift of this reverse process only requires estimates of $Trace\left(\nabla^2 \log p_t\left(\vx_t\right) \right)$, which can be efficiently computed through random projections of the score $\nabla\log p_t(\vx_t)$ that are readily available in generative models through the denoising score matching objective~\citep{hyvarinen2005estimation,dsm}. 
Tweedie's first- and second-order moments are used to estimate the mean and the covariance of the Gaussian used to approximate $p_t(X_0|X_t)$. 
First-order methods approximate $p_t(X_0|X_t)\approx \delta\left(X_0 - \bar{X}_0\right)$, where $\bar{X}_0 = \E_{X_0 \sim p_t\left( X_0|X_t\right)} \left[X_0\right]$. 
Our sampler STSL takes a step beyond the first-order Tweedie, opening plethora of opportunities as a drop-in replacement of traditional first-order samplers~\citep{dps,psld,p2l}. 
Empirically, as our results show, we solve inverse problems in about 50 diffusion steps, compared to 1000 steps in SoTA solvers~\citep{psld,p2l}. 
This translates into 4X and 8X improvement in terms of Neural Function Evaluations (NFEs) over PSLD~\citep{psld} and P2L~\citep{p2l}, respectively. 
Empirically, we demonstrate superior performance in denosing, inpainting, super-resolution, Gaussian deblurring, and motion deblurring tasks on standard benchmarks: FFHQ~\citep{ffhq}, ImageNet~\citep{imagenet}, and COCO~\citep{coco}.

Image editing can be regarded as another form of sampling from the posterior, specifically $p_0(X_0|\vy)$ where $\vy$ is a given image and the goal is to find an edited image $X_0\sim p_0(X_0|\vy)$.
Existing methods either fine-tune generative models for specific tasks~\citep{diffuseCLIP,ruiz2023dreambooth} or use a single foundation model for all tasks~\citep{p2p,p2pZero,nti}, with the latter preferred due to cost and bias concerns~\citep{ruiz2023hyperdreambooth}.
State-of-the-art methods, such as NTI \citep{nti}, excel with clean source images but struggle with real-world corruptions, as seen in Figure \ref{fig:main}.
SoTA solvers like PSLD \cite{psld} and P2L \cite{p2l} can remove this corruption prior to editing, but demand around 1000 diffusion steps to sample $p_0(X_0|\vy)$, making them impractical for editing tasks.
To address this, we repurpose STSL in a two-stage design: first restore the image using our inverse problem solver in just $\sim$50 diffusion steps, and then guide the reverse process in text-based editing using a contrastive loss in feature space.
As we show in our results (\S\ref{sec:exps}), our method overcomes the challenges posed by impractical diffusion step requirements and the content-style preservation issue faced by the SoTA NTI \citep{nti}, and outperforms it in text-guided image editing from corrupted images.

\noindent\textbf{Our contributions are summarized in three-fold:}
\begin{itemize}
    \item We present an efficient second-order approximation using Tweedie's formula to mitigate the bias incurred in the widely used first-order samplers.
    With this method, we devise a surrogate loss function to refine the reverse process at every diffusion step to address inverse problems.
    
    \item We introduce a new framework for high-fidelity image editing in real-world environments with corruptions. 
    To the best of our knowledge, this is the first framework that can handle corruptions in image editing pipelines.
    
    \item We conduct extensive experiments to demonstrate superior performance in tackling inverse problems (such as denoising, inpainting, super-resolution, and deblurring) and achiveing high-fidelity text-guided image editing.
\end{itemize}

\section{Related Work}
\label{sec:rel-work}

\textbf{Inverse Problems:} 
Generative models based on diffusion are gaining popularity as effective priors for solving inverse problems. 
These models fall into two main categories: 
(1) pixel-space diffusion models (PDMs)~\citep{sohl-dickstein15,ddpm,songscore} and (2) latent-space diffusion models (LDMs)~\citep{ldm}. 
Over the years, inverse problem solvers based on PDMs have shown remarkable performance in terms of both quality \citep{pGDM,dps,lugmayr2022repaint,rout2023theoretical,chung2022improving,chung2023direct} and robustness \citep{daras2022soft,mri_paper,bansal2022cold,ajil_posterior_sampling,ddnm}. 
However, these solvers
struggle to generalize effectively to different domains and require a different generative model for each dataset. 
To overcome these limitations, a recent development termed PSLD \citep{psld} leverages the generative power of large foundation models, such as Stable Diffusion. 
PSLD outperforms PDM-based solvers on various datasets by employing a single LDM for all inverse tasks.

PSLD's core concept is
the use of the first-order Tweedie
in the latent space of Stable Diffusion, \ie, $ \log p_{T-t}(\vy|Z_t)\approx \log p_{T-t}(\vy|\gD\left(\E\left[ Z_T|Z_t\right]\right))$. 
To further improve the results, the latents are updated using additional gradients from a gluing objective~\citep{psld}. 
P2L \citep{p2l} extends this idea, and updates both the latents and the text-embeddings along with a generalized version of the gluing objective. 
Both PSLD \citep{psld} and P2L \citep{p2l} rely on gradient-based guidance and require a considerable amount of diffusion steps (around 1000) for reasonable reconstruction. 
In this paper, we show faithful reconstruction with significantly fewer steps (around 50), offering a more practical and efficient approach.

\noindent\textbf{Image Editing:} 
As large foundation models become increasingly accessible, the realm of high-fidelity image editing emerges as a captivating domain for research. 
Similar to inverse problem solvers, image editing tools can be broadly classified into either PDM-based \citep{diffuseCLIP,diffuseIT, ruiz2023dreambooth} or LDM-based \citep{p2p,nti,p2pZero}.
The former requires additional losses, such as CLIP direction loss \cite{diffuseCLIP}, identity loss \cite{diffuseCLIP}, structural similarity loss \cite{diffuseIT}, semantic loss \cite{diffuseIT}, regularization loss \cite{ruiz2023dreambooth}, and face preservation loss \cite{diffuseIT} to guide the reverse process in the pixel space.
On the other hand, LDM-based tools~\citep{p2p,nti,p2pZero} streamline the process by eliminating unnecessary complexities associated with multiple loss functions.
Instead, they leverage cross-attention-control \citep{p2p} on top of a text-conditional generative foundation model. 
However, as illustrated in Figure \ref{fig:main}, these methods fail to apply faithful edits when confronted with real-world corruptions. 
Moreover, the edits are not consistently localized in the absence of corruptions. 
We pinpoint the fundamental cause of this failure and introduce a framework designed to address such real-world corruptions (\S\ref{sec:exps}).

\noindent\textbf{Second-order Correction to the Tweedie Estimator:} 
The first-order Tweedie estimator in Eq.~\eqref{eq-mean-approx} plays a pivotal role in both inverse problem solvers \citep{dps,psld,p2l} and image editing tools \citep{diffuseCLIP,diffuseIT}. 
However, this estimator is biased towards $\E_{Z_T\sim P(Z_T|Z_t)} \left[Z_T\right]$ rather than generating samples $Z_T \sim p_{T-t}(Z_T|Z_t)$, resulting in reconstruction lacking sharpness and details, as shown in \S\ref{sec:exps}. 
This bias is attributed to the significant Jensen's gap \cite{dps}, a concept we delve into in  \S\ref{sec:theory}.
To address this limitation, we propose an alternative Tweedie estimator in \S\ref{sec:method}.
Notably, our method requires only first-order scores, unlike prior methods \cite{tmpd,meng2021estimating} that require second-order score or the Jacobian of the first-order score. 
By focusing on the first order score $\nabla \log p_{T-t}(\vy|Z_t)$ in sampling $p_0(X_0|\vy)$, our estimator proves to be a seamless substitute for the Tweedie estimator (\ref{eq-mean-approx}) with minimal computational overhead.
As shown in \S\ref{sec:exps}, our proposed estimator has a profound impact across various tasks, including inpainting, super-resolution, deblurring, denoising
, and text-guided image editing, owing to the crucial role it plays in sampling $p_0(X_0|\vy)$.

\vspace{-0.5ex}
\section{Method}
\vspace{-1ex}
\label{sec:method}

\subsection{Background}
\vspace{-0.5ex}
\label{sec:background}
\textbf{Diffusion Probabilistic Models (DPMs):} 
 DPMs \cite{sohl-dickstein15,ncsn,ddpm} consist of two stochastic processes, the forward process and the reverse process.
In the forward process, noise is gradually added to a clean image until it becomes indistinguishable from pure noise. 
This progression is characterized by the general It\^{o} Stochastic Differential Equation (SDE): $dX_t = b(X_t, t) dt + \sigma(X_t, t)dW_t$,
where $X_t\in \R^{d}$, drift $b:\R^{d}\times \R_+ \rightarrow \R^d$, volatility $\sigma:\R^{d}\times \R_+ \rightarrow \R$, and $\{W_t\}_{t=0}^T$ is an $n$-dimensional Wiener process (or Brownian motion) \cite{oksendal2013stochastic}. 
In practice, the forward process is commonly implemented as an Ornstein-Uhlenbeck (OU) process:
\begin{equation}
    \label{eq-forward}
    dX_t = -X_t dt + \sqrt{2}dW_t,
\end{equation}
which has a solution of $X_t = X_0 e^{-t} + \sqrt{2} \int_0^t e^{-(t-s)}dW_s$ that induces a Gaussian transition kernel as given by $p_t\left(X_t|X_0=\vx_0\right) = \gN\left(X_t; \vx_0 e^{-t}, (1-e^{-2t}) I \right)$. 
In a discrete setting, this can be written as $p_t\left(X_t|X_0=\vx_0\right) = \gN\left(X_t; \sqrt{\bar{\alpha}_t}\vx_0, (1-\bar{\alpha}_t) I\right) $, where $\bar{\alpha}_t = \prod_{s=0}^t \alpha_s$ for a finite sequence of $\alpha_s\in[0,1]~\forall s \in [0,T]$. 

On the other hand, the reverse process gradually removes noises 
to produce a clean sample at the end, as characterized by the reverse It\^{o} SDE: $dZ_t = \left(Z_t + 2 \nabla \log p_{T-t} (Z_t) \right)  dt + \sqrt{2} d\tilde{W}_t$, subject to weak regularity conditions \citep{anderson}.
To achieve this, a neural network is trained to approximate the score $\nabla \log p_{T-t}(Z_t) \approx s_\vtheta(Z_t,T-t)\, \forall t\in[0,T]$ \citep{dsm,hyvarinen2005estimation}, 
and the reverse It\^{o} SDE: $dZ_t = \left(Z_t + 2 s_\vtheta(Z_t,T-t) \right)  dt + \sqrt{2} d\tilde{W}_t$ is employed to sample from the data distribution $p_{data}(X)\coloneqq p_0(X_0)$.

\textbf{Posterior Sampling:} 
In this regime, the objective is to sample from $p_0(X_0|\vy)$ that leads to a conditional It\^{o} SDE: $dZ_t \!=\! \left(Z_t + 2 \nabla \log p_{T-t} (Z_t|\vy) \right)  dt + \sqrt{2}d\tilde{W}_t$.
Using Bayes' theorem, the drift term breaks down into $\left(Z_t + 2 \nabla \log p_{T-t} (\vy|Z_t)+2\nabla \log p_{T-t} (Z_t) \right)$. 
Generative foundation models, such as Stable Diffusion \citep{ldm}, Imagen \citep{saharia2022photorealistic}, and DALL-E \citep{ramesh2021zero,ramesh2022hierarchical} offer a reliable approximation of the true score, \ie, $s_\vtheta(Z_t,T-t) \approx \nabla \log p_{T-t}(Z_t)$~\citep{dsm,hyvarinen2005estimation}. 
As a result, recent focus has shifted towards approximating $ \log p_{T-t}(\vy|Z_t)$. 
In the context of solving inverse problems or editing natural images with specific prompts, an interesting line of research \citep{dps,diffuseCLIP,psld,p2l,diffuseIT} approximates
\begin{equation}
    \label{eq-mean-approx}
    \log p_{T-t}(\vy|Z_t)\approx \log p_{T-t}(\vy|\E_{p_{T-t}(Z_T|Z_t)} \left[Z_T\right]).
\end{equation} 
We refer to (\ref{eq-mean-approx}) as the first-order Tweedie estimator for Pixel-space Diffusion Models (PDMs).
For Latent-space Diffusion models (LDMs), PSLD~\citep{psld} proposes the following first-order approximation (labeled as LDPS in \S\ref{sec:exps}): 
\begin{equation}
    \label{eq-psld-mean-approx}
    \log p_{T-t}(\vy|Z_t)\approx \log p_{T-t}(\vy|\dec(\E_{p_{T-t}(Z_T|Z_t)} \left[Z_T\right])),
\end{equation} 
where $\dec(.)$ denotes a decoder from latent to pixel space. We will denote the pixel to latent encoder by $\enc(.)$.

\subsection{STSL for Image Inversion}
\label{sec:algo-inv}
\begin{algorithm}[!t]
    \scriptsize
    \caption{STSL for Image Inversion}
    \label{alg:stsl}
         \KwIn{
         diffusion time steps $T$,
         observed $\vy$,
         measurement operator $\mA$,
         likelihood strength $\lambda$,
         stochastic averaging steps $K$,
         second-order correction stepsize $\eta$,
         encoder $\enc$,
         decoder $\dec$,    
         learned score  $\bm{s}_\theta
         $
         }
          Initialization: $\overrightarrow{Z}_0 = \enc(\mA^T\vy)$ \hfill $\triangleright$ forward process\\
          \For{$t=0$ \KwTo $T-1$}{
          $\overrightarrow{Z}_{t+1}
          \gets 
          \sqrt{\alpha_{t}}\overrightarrow{Z}_{t}
          -\sqrt{(1-\alpha_t)} \sqrt{(1-\bar{\alpha}_t)} s_\vtheta(\overrightarrow{Z}_{t}, t) $\\
          }
          Initialization: $Z_0 = \overrightarrow{Z}_T$ \hfill $\triangleright$ reverse process\\
          \For{$t=0$ \KwTo $T-1$}{
            \For{$k=0$ \KwTo $K$}{
                $\veps \sim \gN(\bm{0}, \mI)$ \hfill $\triangleright$ stochastic averaging\\
                $\bar{Z}_T
                \gets 
                (Z_t + (1 - \bar\alpha_{T-t}) \bm{s}_\vtheta(Z_t, T-t))/ \sqrt{\bar\alpha_{T-t}}
                $\\
                $Z_{t} \gets Z_{t} - \lambda \nabla \lVert \vy - \mA\dec(\bar{Z}_T)\rVert
                $ 
                \hfill$- 
                (\eta/d)
                  \nabla \left(\veps^T \bm{s}_\vtheta\left(Z_t+ \veps,T-t\right)
                  - \veps^T\bm{s}_\vtheta\left(Z_t,T-t \right)
                  \right)$\\
            }
            $\bar{Z}_T
            \gets 
            (Z_t + (1 - \bar\alpha_{T-t}) \bm{s}_\vtheta(Z_t, T-t))/ \sqrt{\bar\alpha_{T-t}}
            $\\
            $Z_{t+1} 
            \gets
            \frac{\sqrt{\alpha_{T-t}}(1-\bar{\alpha}_{T-t-1})}{1 - \bar{\alpha}_{T-t}}Z_t
            + \frac{\sqrt{\bar{\alpha}_{T-t-1}}(1-\alpha_{T-t})}{1 -\bar{\alpha}_{T-t}}\bar{Z}_T 
            $\\
            }
            \Return $\dec(Z_T)$
\end{algorithm}

The proposed Algorithm~\ref{alg:stsl},  
termed STSL, differs from prior methods \citep{psld,p2l,nti,p2p} in two key aspects: initialization and latent refinement.
We discuss each in turn.

\textbf{Initialization:}
Existing solvers \citep{psld,p2l} initiate the reverse process from $Z_0\sim \pi_d$, and incur a discretization error of $\gO \left(de^{-2T}\right)$ that comes from $D_{KL}\left(p_T||\pi_d\right)$ \citep{chen2022sampling,benton2023linear}. 
As we aim to sample $p_0(X_0|\vy)$ with \textit{fewer diffusion steps}, this error can be substantial in \textit{high-dimensional} sampling. 
To address this, we reduce the error by initializing the reverse process at $Z_0\sim p_T(Z_0|\enc(\mA^T\vy))$, and run the forward process in Eq.~\eqref{eq-forward} using the Gaussian transition kernel, starting from $\overrightarrow{Z_0} = \enc(\mA^T\vy)$ (\S\ref{sec:background})\footnote{An alternative could be to initialize $\overrightarrow{Z_0}$ as $\arg\min_Z \lVert \vy-\mA \mathcal{D}(Z)\rVert_2^2$. However, this is more expensive. Empirically, we observe good results using one step projection, which is significantly less expensive.}. 
The final latent $\overrightarrow{Z_T} \stackrel{\pi_d}{=} Z_0$ (equal in distribution) is then employed for initialization.

\textbf{Refinement:} With $Z_0\sim p_T(Z_0|\enc(\mA^T\vy))$, we propose to sample from a new reverse It\^{o} SDE as follows:
\begin{equation}
    \label{eq-reverse-posterior-refine}
    dZ_t = \tilde{b}(Z_t,t)  dt + \sqrt{2} d\tilde{W}_t,
\end{equation}
where $\tilde{b}(Z_t,t) \coloneqq \left(Z_t + \gG(\vy,Z_t)  +2\nabla \log p_{T-t} (Z_t) \right)$ is the new drift. 
In prior works \citep{dps,psld,p2l}, $\gG(\vy,Z_t)$ represents a single gradient of the log likelihood evaluated at $\bar{Z}_T$, \ie $\gG(\vy,Z_t)\approx \nabla \log p_{T-t}(\vy|\bar{Z}_T) = -\lambda \nabla \lVert\vy-\mA\dec(\bar{Z}_T) \rVert_2^2$, where $\bar{Z}_T \!\coloneqq\! \E_{Z_T\sim p_{T-t}(Z_T|Z_t)}\left[Z_T\right] \!=\! \frac{Z_t}{\sqrt{\bar{\alpha}_t}} \!+\! \frac{(1-\bar{\alpha}_t)\nabla \log p_{T-t}(Z_t)}{\sqrt{\bar{\alpha}_t}}$. 
This correction step, essential for SoTA solvers~\cite{psld,p2l} after each denoising update, introduces a quality-limiting bias due to the regression to the mean $\bar{Z}_T$, as illustrated in \S\ref{sec:exps}.

In contrast, we propose an update rule that considers a second order correction term using a surrogate loss function alongside a proximal gradient update.
Specifically, we rewrite $\gG(\vy,Z_t)$ by approximating it as $-\nabla \gL(\vy,Z_t)$, with $\gL(\cdot,\cdot)$ denoting the surrogate loss function:
\begin{align}
    &
    \gL(\vy,Z_t) 
    \coloneqq
      \lambda \lVert\vy-\mA\dec(\bar{Z}_T) \rVert_2^2 
      \label{eq-surrogate-loss}
      \\
      \nonumber
    & +
      \frac{\eta}{d} \E_{\veps \sim \pi_d}\left[ \veps^T \left(\nabla \log p_{T-t}\left(Z_t+ \veps\right) - \nabla \log p_{T-t}\left(Z_t\right)\right)\right],
\end{align}
justified theoretically in \S\ref{sec:theory}.
To ensure we recover $X_0$ at the end of this new reverse process in Eq.~\eqref{eq-reverse-posterior-refine}, we optimize for $Z_t$ within a small neighborhood around the corresponding forward latent $\overrightarrow{Z}_{T-t}$, which was sampled during the {\em Initialization} step outlined above. 
This is equivalent to an iterative proximal gradient update using $\gL(\vy,Z_t) + \kappa \lVert Z_t - \overrightarrow{Z}_{T-t}\rVert_2^2$ for some $\kappa > 0$. 

In our implementation, instead of solving a local optimization, we opt for a simpler strategy by iterative gradient updates using $\gL(\vy,Z_t)$, and replacing the expectation in Eq.~\eqref{eq-surrogate-loss} with a single-sample estimate at each step, \ie, draw a random $\veps \!\!\sim\!\! \pi_d$ to perform the update. 
This iterative process of local updates accumulates into an estimate of the expectation due to path-wise stochastic averaging, which has proven effective in online learning~\citep{lattimore2020bandit} and stochastic approximation~\cite{borkar08}.
Further, we add contrastive loss to the surrogate loss as $\gL(\vy,Z_t) + (\nu/d) \gL_{ViT}\left( \vy, \mA\dec(\bar{Z}_T)\right)$~\citep{diffuseIT} to improve the perceptual quality (\S\ref{sec:ablation}).
For simplicity, we do not include this term in Algorithm~\ref{alg:stsl}.
This new reverse process is now tractable, and exhibits performance improvements over SoTA solvers. 
Ultimately, the solution to inversion materializes as $\dec(Z_T)$, obtained at the end of Eq.~\eqref{eq-reverse-posterior-refine}.
\vspace{-5ex}
\subsection{STSL for Image Editing}
\vspace{-2ex}
\label{sec:algo-edit}
To edit a real image, NTI \citep{nti} stands out as a leading method that associates the real image with a sequence of null embeddings.
Formally, define $\Phi(\cdot)$ as an encoder that maps a text prompt to an embedding in $\R^h$. 
Given a text-conditional score network $s_\vtheta: \R^d \!\times\! \R_+ \!\times\! \R_h \!\rightarrow\! \R^d$, NTI \citep{nti} tackles the optimization problem $\hat{\varphi}_{t} = \arg \min_{\varphi_{t}} \lVert\overrightarrow{Z}_{T-t-1} - f\left(Z_t,T-t,\varphi_{t} \right) \rVert_2^2$,
with $\{\varphi_t\}_{t=0}^T$ initialized by null-text embeddings 
and $f\left(Z_t,T-t,\varphi_{t} \right) \!=\! \sqrt{\bar{\alpha}_{T-t-1}}\bar{Z}_T \!+\! \sqrt{1-\bar{\alpha}_{T-t-1}}\sqrt{1-\bar{\alpha}_{T-t}}s_\vtheta(Z_t,T \!-\! t,\varphi_{T-t})$.
Instead, we propose to solve: $\hat{\varphi}_{t} = \arg \min_{\varphi_{t}} \lVert Z_{t+1} - f\left(Z_t,T-t,\varphi_{t} \right) \rVert_2^2$, where $\{Z_t\}_{t=0}^T$ are obtained from our novel reverse SDE using the surrogate loss in Eq.~\eqref{eq-surrogate-loss}.
NTI \citep{nti} associates $\{\varphi_t\}_{t=0}^T$ with the corrupt image, leaving residual corruptions in the edited image (see \S\ref{sec:exps-editing}).
In contrast, our proposed null-optimization aligns $\{\varphi_t\}_{t=0}^T$ with a clean image because our new SDE in Eq.~\eqref{eq-reverse-posterior-refine} yields a clean image at the end.
Thereby, it enables text-guided noisy image editing via Cross-Attention-Control (CAC)~\citep{p2p}.

Conventional CAC-based image editing \citep{p2p} encounters a critical limitation -- it struggles to maintain the original image content while incorporating the desired modifications (\S\ref{sec:exps-editing}).
To address this issue, our method refines the latents after CAC update via posterior sampling. 
Let $\gC$ denote the CAC module that gives $\hat{Z}_{t+1} = \gC\left(Z_t, \Phi(\text{``prompt"}),Z_t,\hat{\varphi}_{t}\right).$
Then, we update $Z_{t+1}$ using a single step of STSL:

\begin{align}
    & 
    Z_{t+1} 
    \leftarrow \hat{Z}_{t+1} - 
    \lambda \nabla \lVert\vy-\mA\dec(\bar{Z}_T) \rVert_2^2 
    \label{eq-edit-mse}
    \\
    \nonumber
    &
    -
    \frac{\eta}{d} \nabla \E\left[ \veps^T \left(\nabla \log p_{T-t}\left(Z_{t+1}+ \veps\right) - \nabla \log p_{T-t}\left(Z_{t+1}\right)\right)\right].
\end{align}
Inverse problem solvers and image editing tools have distinct differences, mainly because image editing tools lack groundtruth measurements.
This reduces the usefulness of the measurement update $\nabla \lVert\vy-\mA\dec(\bar{Z}_T) \rVert_2^2$ during editing. 
To address this, we employ a contrastive loss between the edited image $\dec(\bar{Z}_T)$ and the input image in the feature space, with features obtained through ViT's multi-head self-attention layers \citep{tumanyan2022splicing}. 
However, using ViT for feature extraction in early-stage diffusion could yield undesired results due to its inherent focus on natural image domain. 
To ensure meaningful features, we apply measurement updates in the early stage (around 30 steps) of the reverse process, and then update the latents in Eq.~\eqref{eq-edit-mse} using the contrastive loss \citep{diffuseIT}: $Z_{t+1}\leftarrow Z_{t+1} - \frac{\nu}{d} \nabla \gL_{ViT}\left( \vy, \mA\dec(\bar{Z}_T)\right).$
This keeps the image's \textit{content} while infusing desired \textit{semantics} via the cross-attention features. 
We term this process Cross-Attention-Tuning (CAT), and our proposed image editing method STSL-CAT.

\section{Theory}
\label{sec:theory}

Recall from \S\ref{sec:background} that we seek for a good approximation of $\log p_{T-t}(\vy|Z_t)$. 
Methods based on Tweedie's first order moment leverage the posterior mean from \textbf{Proposition~\ref{prop-tweedie}} to approximate $\nabla \log p_{T-t}(\vy|Z_t) \!\approx\! \nabla \log p_{T-t}(\vy|\bar{Z}_T)$, where $\bar{Z}_T = \E_{Z_T \sim p_{T-t}\left( Z_T|Z_t\right)}[Z_T]$. 
Since $s_\vtheta(Z_t,T-t) \approx \nabla \log p_{T-t}(Z_t)$, this enables a practical implementation of posterior samplers \citep{dps,pGDM,psld,p2l}.

\begin{proposition}[\citep{robbins1956empirical,efron2011tweedie}]
    \label{prop-tweedie}
    Given $\vx_t \!\!=\!\! \sqrt{\bar{\alpha}_t}\vx_0 \!\!+\!\! \sqrt{1-\bar{\alpha}_t}\mathbf{\veps}$ and $\veps \!\!\sim\!\! \gN\left(0, I\right)$, denote by $\bar{X}_0 \!\!=\!\! \E_{X_0 \sim p_{t}\left( X_0|X_t=\vx_t\right)}[X_0]$ the posterior mean of $p_t(X_0|X_t \!=\! \vx_t)$. 
    Then, for the variance preserving SDE in \eqref{eq-forward} or DDPM sampling~\citep{ddpm}, $p_t(X_0|X_t=\vx_t)$ has mean $\bar{X}_0 \!=\! \frac{\vx_t}{\sqrt{\bar{\alpha}_t}}+\frac{(1-\bar{\alpha}_t)}{\sqrt{\bar{\alpha}_t}}\nabla_{\vx_t} \log p_t(X_t=\vx_t)$ and covariance $\E_{X_0 \sim p_t(X_0|X_t=\vx_t)}\left[\left(X_0 \!-\! \bar{X}_0\right)\left(X_0 \!-\! \bar{X}_0\right)^T\right] \!\!=\!\! \frac{1-\bar{\alpha}_t}{\bar{\alpha}_t}\left(\mI + (1-\bar{\alpha}_t)\nabla_{\vx_t}^2 \log p_t(X_t=\vx_t)\right)$.
    
\end{proposition}
The error from the first-order approximation is characterized by the Jensen's gap [\citealp[Theorem 1]{dps}]. 
The gap can be notable in practice due to the local linearity of first-order approximations \citep{meng2021estimating}.
Prior works \citep{meng2021estimating,tmpd} mitigate this error by a second-order approximation that introduces curvature to the estimator.
However, these methods \citep{meng2021estimating,tmpd} require expensive Hessian computations. 
In contrast, our sampler requires only the \textit{trace} of the Hessian, and hence enable efficient computation with minimal cost.
We present our main results in \textbf{Theorem~\ref{thm:unbiased-tweedie}} under the assumptions stated below.
\begin{assumption}
    \label{assm-meas-exp}
    Define $\bar{Z}_T \coloneqq \E_{Z_T \sim p_{T-t}\left( Z_T|Z_t\right)}[Z_T]$ for $t\in [0,T]$. Then,     
     $p_{T-t}(\vy|\bar{Z}_T)=\gN\left(\vy;\mA\bar{Z}_T,\sigma_{\vy}^2\mI\right)$ and  $p_{T-t}(\vy|\bar{Z}_T) > 0 $.
\end{assumption}

\begin{assumption}
    \label{assm-meas-func}
     For all $\tilde{Z}_T\in \R^d$ and $m>0$, $-m p_{T-t}(\vy|\bar{Z}_T) \mI \preceq \nabla^2 p_{T-t}(\vy|Z_t)\big|_{\tilde{Z}_T}$, where $\bar{Z}_T \coloneqq \E_{Z_T \sim p_{T-t}\left( Z_T|Z_t\right)}[Z_T]$ for $t \in [0,T]$.
\end{assumption}

\textbf{Assumption \ref{assm-meas-exp}} is a common condition in prior works \citep{dps,rout2023theoretical,psld,p2l}, indicating informative measurements from $\bar{Z}_T$. 
{\color{black} \textbf{Assumption~\ref{assm-meas-func}} simplifies mathematical considerations and ensures the smallest eigenvalue of the Hessian
is uniformly lower bounded by a finite quantity.} 
Notably, the widely used Gaussian measurement model $\vy=\mA\bar{Z}_T + \sigma_\vy^2 \vn, \vn\sim \gN\left(\vzero,\mI \right)$ satisfies both these assumptions.

\begin{theorem}[Tweedie Sampler from Surrogate Loss]
    \label{thm:unbiased-tweedie}
    Suppose \textbf{Assumption~\ref{assm-meas-exp}} and \textbf{Assumption~\ref{assm-meas-func}} hold.
    Let $\hat{\gL}(\vy,Z_t) \coloneqq \log \left( p_{T-t}(\vy|\bar{Z}_T) \right)
    + \log \left( \xi_t - (1-\bar{\alpha}_t) m~Trace\left(\nabla^2 \log p_{T-t}(Z_t) \right)
    \right)$,
    where $\xi_t = 
    1-
    \frac{1-\bar{\alpha}_t}{\bar{\alpha}_t} md
    .$ For $\lambda=\gO(\frac{1}{\sigma_\vy^2})$ and $\gamma = \gO(\frac{\eta}{d})$, the following holds:
    $\hat{\gL}(\vy,Z_t) \leq \log p_{T-t}\left(\vy|Z_t\right)$
    and the gradient of $\hat{\gL}(\vy,Z_t)$ becomes
    $\nabla \hat{\gL}(\vy,Z_t) \simeq - \lambda \nabla \lVert\vy-\mA\bar{Z}_T \rVert_2^2 - \gamma \nabla \left(Trace\left(\nabla^2 \log p_{T-t}(Z_t) \right) \right)$.
\end{theorem}

\begin{proof}
    The proof is included in Appendix~\ref{sec:prf-tsl}.
\end{proof}

We draw the following insights from \textbf{Theorem~\ref{thm:unbiased-tweedie}.}
\textbf{Practically Implementable:} Given $\veps \!\!\sim\!\! \gN\left( \vzero, \mI \right)$, we use Hutchinson's estimator~\citep{hutchinson1989stochastic}, along with a random projection based gradient estimator \cite{flkamc05} to compute the trace of the Hessian as given below 
$Trace\left(\nabla^2 \log p_{T-t}(Z_t) \right) \!\simeq\! \E\left[ \veps^T \left(\nabla \log p_{T-t}\left(Z_t  + \veps\right) \!-\! \nabla \log p_{T-t}\left(Z_t\right)\right) \right] \!-\! \gO(\lVert\veps\rVert^3)$ 
(Appendix~\ref{sec:prf-hutchinson} for details).
  
Substituting this in \textbf{Theorem~\ref{thm:unbiased-tweedie}} and extending the result to LDMs, our algorithm becomes tractable because it requires $\nabla \log p_{T-t}(Z_t) \!\approx\! s_\theta(Z_t,T-t)$, which is readily available in LDMs~\citep{ldm}:
\begin{align}
    &
    \nabla \hat{\gL}(\vy,Z_t)
    \simeq
    -
    \lambda \nabla \lVert\vy-\mA\dec(\bar{Z}_T) \rVert^2 
    \label{eq-unbiased-ldm}
    \\
    \nonumber
    &
    -
    \gamma \nabla \E_{\veps\sim\pi_d}\left[ \veps^T \left(\nabla \log p_{T-t}\left(Z_t  + \veps\right) - \nabla \log p_{T-t}\left(Z_t\right)\right)
        \right].
\end{align}
Further, $\nabla \hat{\gL}(\vy,Z_t) \simeq -\nabla \gL(\vy,Z_t)$, where $\gL(\vy,Z_t)$ is the surrogate loss function in \S\ref{sec:algo-inv} (see Appendix~\ref{sec:prf-tsl} for details).

\noindent\textbf{Connection with First-order Tweedie:} 
The update in Eq.~\eqref{eq-unbiased-ldm} samples from an alternate reverse process
$dZ_t = \tilde{b}(Z_t,t)  dt + \sqrt{2} d\tilde{W}_t$.
Thus, in Eq.~\eqref{eq-mean-approx}, if we modify the first-order Tweedie's conditional drift by setting $\gamma=0$ and applying a one-step gradient of $\gL(\vy,Z_t)$, it becomes a special case of our drift $\tilde{b}(Z_t,t)$. 
However, this setup introduces a bias that could hamper the quality, as we illustrate in \S\ref{sec:exps}. 
To mitigate this bias, we implement multiple proximal gradient updates and leverage stochastic averaging to estimate expectations using the surrogate loss $\gL\left(\vy,Z_t\right)$.

\begin{table*}[!thb]
\centering
\setlength{\tabcolsep}{2pt}
\resizebox{\textwidth}{!}{%
\begin{tabular}{lcccccccccccr}
\toprule
{} & \multicolumn{3}{c}{\textbf{SR ($\times 8$)}} & \multicolumn{3}{c}{\textbf{Motion Deblur}} & \multicolumn{3}{c}{\textbf{Gaussian Deblur}} 
\\
\cmidrule(lr){2-4}
\cmidrule(lr){5-7}
\cmidrule(lr){8-10}
{\textbf{Method}} & {LPIPS$\downarrow$} & {PSNR$\uparrow$} & {SSIM$\uparrow$} & {LPIPS$\downarrow$} & {PSNR$\uparrow$} & {SSIM$\uparrow$} & {LPIPS$\downarrow$} & {PSNR$\uparrow$} & {SSIM$\uparrow$}\\
\midrule
\rowcolor{Orange!10}
STSL~(Ours) & \textbf{0.335} & 31.77 & 91.32 & \textbf{0.321} & \textbf{31.71} & \textbf{90.01} & \textbf{0.308} & \underline{32.30} & \textbf{94.04} \\
\cmidrule(l){1-10}
P2L~\citep{p2l} & 0.381 & 31.36 & 89.14 & \underline{0.395} & 31.37 & \underline{88.81} & 0.382 & 31.63 & 90.89  \\
PSLD~\citep{psld} & 0.402 & 31.39 & 88.89 & 0.408 & 31.37 & 87.61 & 0.371 & 32.26 & 92.63 \\
 GML-DPS~\citep{psld} & 0.368 & \underline{32.34} & \underline{92.69} & 0.408 & \underline{31.43} & 87.99 & \underline{0.370} & \textbf{32.33} & \underline{92.65} \\
LDPS~\citep{psld} & \underline{0.354} & \textbf{32.54} & \textbf{93.46} & 0.433 & 31.21 & 86.29 & 0.385 & 32.18 & 92.34 \\
LDIR~\citep{he2023iterative} & 0.423 & 30.98 & 86.77 & 0.446 & 30.33 & 80.05 & 0.421 & 30.78 & 84.65 \\
\cmidrule(l){1-10}
\rowcolor{gray!10}
DPS~\citep{dps} & 0.538 & 29.15 & 72.92 & 0.556 & 28.98 & 71.95 & 0.694 & 28.14 & 56.15\\
\rowcolor{gray!10}
DiffPIR~\citep{zhu2023denoising} & 0.791 & 28.12 & 42.62 & 0.593 & 28.92 & 67.09 & 0.614 & 29.06 & 73.51 \\
\bottomrule
\end{tabular}
\hspace{0.01cm}
\begin{tabular}{lccccccccccc}
\toprule
\multicolumn{3}{c}{\textbf{SR ($\times 8$)}} & \multicolumn{3}{c}{\textbf{Motion Deblur}} & \multicolumn{3}{c}{\textbf{Gaussian Deblur}} 
\\
\cmidrule(lr){1-3}
\cmidrule(lr){4-6}
\cmidrule(lr){7-9}
{LPIPS$\downarrow$} & {PSNR$\uparrow$} & {SSIM$\uparrow$} & {LPIPS$\downarrow$} & {PSNR$\uparrow$} & {SSIM$\uparrow$} & {LPIPS$\downarrow$} & {PSNR$\uparrow$} & {SSIM$\uparrow$}\\
\midrule
\rowcolor{Orange!10}
\textbf{0.392} & \textbf{30.64} & \textbf{84.86} & \textbf{0.420} & 30.38 & 76.45 & \textbf{0.349} & 31.03 & \textbf{90.21}\\
\cmidrule(l){1-9}
\underline{0.441} & \underline{30.38} & \underline{84.27} & \underline{0.429} & \textbf{30.65} & \textbf{84.53} & 0.402 & 30.70 & 87.92\\
0.484 & 30.23 & 80.72 & 0.433 & 30.50 & 82.17 & \underline{0.387} & \textbf{31.20} & \underline{88.65} \\
0.482 & 30.35 & 81.01 & 0.432 & \underline{30.55} & \underline{82.65} & \underline{0.387} & \underline{31.19} & 88.60  \\
0.480 & 30.25 & 80.87 & 0.462 & 30.36 & 80.03 & 0.402 & 31.14 & 88.19 \\
0.498 & 30.04 & 78.22 & 0.531 & 29.56 & 68.32 & 0.501 & 29.81 & 74.56  \\
\cmidrule(l){1-9}
\rowcolor{gray!10}
0.522 & 29.18 & 70.72 & 0.548 & 28.79 & 68.86 & 0.647 & 28.11 & 55.40 \\
\rowcolor{gray!10}
0.740 & 28.17 & 42.80 & 0.577 & 28.81 & 61.20 & 0.612 & 28.93 & 68.67  \\
\bottomrule
\end{tabular}
}
\vspace{-1ex}
\caption{
{\bf Quantitative results of inversion:} On FFHQ-1K (512$\times$512, \textbf{left}) and ImageNet-1K (512$\times$512, \textbf{right}), the results are obtained with a noise level $\sigma_\vy=0.01$. 
Best methods are emphasized in \textbf{bold} and second best \underline{underlined}. 
PDM-based solvers are shaded in gray; LDM-based solvers are in the middle row block.
Notably, our method STSL outperforms leading inverse problem solvers \citep{psld,p2l}.
}
\label{tab:ffhq-imagenet}
\vspace{-1ex}
\end{table*}

\begin{figure*}
    \includegraphics[width=\linewidth]{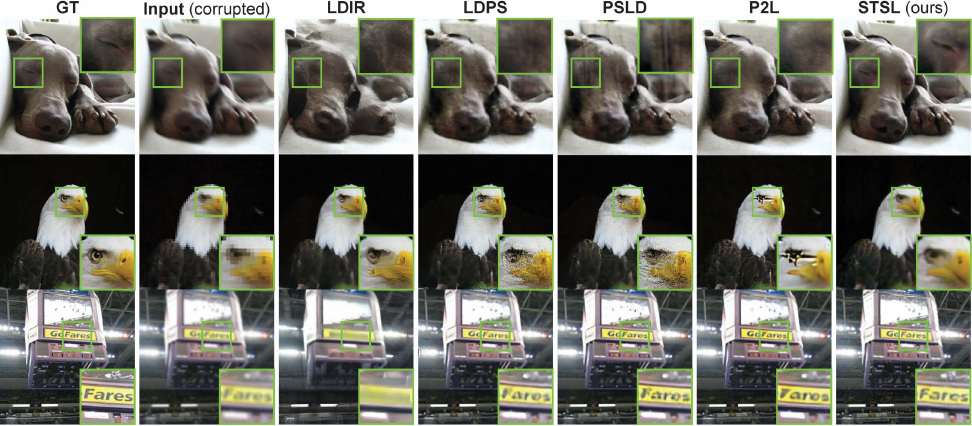}
    \vspace{-4ex}
    \caption{
    \textbf{Qualitative results of inversion on ImageNet (512$\times$512):} 
    All the compared methods utilize the same foundation model, Stable Diffusion. 
    The top, middle, and bottom rows represent the results on Motion Deblur, Super Resolution (8X), and Gaussian Deblur, respectively. 
    The highlighted segment distinctly reveals the superior performance of our method (\textsf{STSL}).
    }
    \vspace{-1ex}
    \label{fig:exp:inversion}
    \vspace{-0.1in}
\end{figure*}

\noindent\textbf{Computational Complexities:} 
Using the gradient update given in Eq.~\eqref{eq-unbiased-ldm}, our sampler provides an efficient second-order approximation. 
It uses a correction step, denoted as $\nabla \E_{\veps\sim\pi_d  }\left[ \veps^T \left(\nabla \log p_{T-t}\left(Z_t+ \veps\right) - \nabla \log p_{T-t}\left(Z_t\right)\right)\right]$, which we estimate using only $\gO(1)$ number of steps.
This stands in contrast to prior methods that require $\gO(d^2)$ compute, such as calculating the Jacobian of the score $\nabla\left(\nabla \log p_{T-t}\left(Z_t\right) \right)$ \citep{tmpd}, or even more expensive approaches like re-training of the Hessian $\nabla^2 \log p_{T-t}(Z_t)$ \citep{meng2021estimating}. 
Further theoretical results are deferred to Appendix~\ref{sec:app-theory}.


\vspace{-0.5ex}
\section{Experiments}
\label{sec:exps}
\vspace{-1.5ex}

\noindent\textbf{Datasets:} 
We adhere to FFHQ \citep{ffhq} and ImageNet \citep{imagenet} benchmarks with 512$\times$512 images. 
For FFHQ, we use the identical set of 1000 images as prior work \cite{dps,psld,p2l}.
For ImageNet, we follow P2L \cite{p2l} by uniformly sampling 1000 images from the {\tt ctest10k} split \citep{saharia2022palette}. 
We conduct ablation studies using COCO 2017 validation set \citep{coco}.

\noindent\textbf{Baselines:} 
For inverse problems, we benchmark against SoTA solvers PSLD~\citep{psld} and P2L~\citep{p2l}, alongside LDM-based methods LDPS \citep{psld}, GML-DPS \citep{psld} and LDIR \citep{he2023iterative}. 
For completeness, we also extend comparisons to PDM-based solvers DPS~\citep{dps} and DiffPIR~\citep{zhu2023denoising}. 
For image editing, we compare with a leading solution NTI~\citep{nti}.

\noindent\textbf{Inverse Tasks:} 
We consider five inverse tasks: motion deblurring, super-resolution (SR), Gaussian deblurring, random inpainting, and denoising.
We follow the setup of P2L \citep{p2l} for motion deblurring\footnote{\url{https://github.com/LeviBorodenko/motionblur}}, Gaussian deblurring, and super-resolution (8X). 
While testing SR at 8X could be ambitious, it provides a challenge to push these algorithms to their limits. 
We also test in less demanding inverse problems of SR at 4X, random inpainting with 40\% dropped pixels, and denoising for salt-pepper noise with 2\% noises.

\noindent\textbf{Metrics:} 
We evaluate using standard metrics: LPIPS, PSNR and SSIM.
For editing, we resort to CLIP accuracy \citep{p2pZero}.
All experiments were conducted on a single A100 GPU.
See Appendix~\ref{sec:impl-details} for details.

\begin{figure*}
    \includegraphics[width=\linewidth]{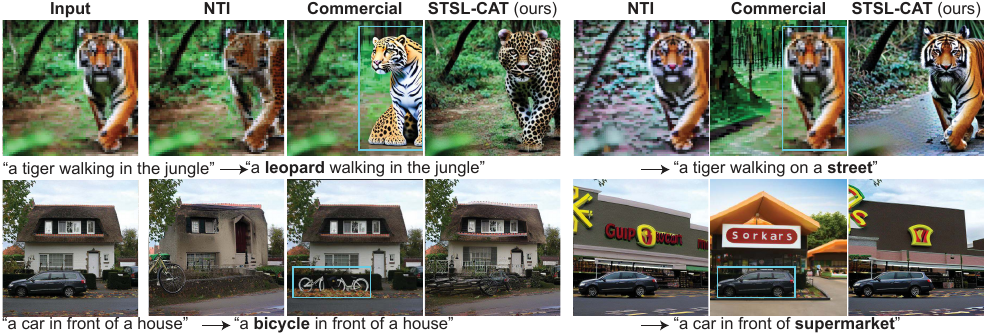}
    \vspace{-4ex}
    \caption{
    {\bf Qualitative results on image editing:}
    We compared our method with NTI \cite{nti} and commercial software on images with corruption (SRx8 {\bf top}) and without ({\bf bottom}).
    The commercial software requires an additional mask to localize the edits (cyan rectangles). 
    }
    \label{fig:exp:edit}
    \vspace{-2ex}
\end{figure*}

\begin{table}[!t]
 \small
  \begin{tabu} to \linewidth {@{}l@{}ccc@{}}
    \toprule
    \textbf{Method} & \textbf{Runtime(s)} & \textbf{NFEs} & \textbf{Steps } \\
    \midrule
    \rowcolor{orange!10}
    STSL (ours) & {\color{black}45} & 250 & 50 \\
    P2L~\citep{p2l} & {\color{black}500} & 2000& 1000 \\
    PSLD~\citep{psld} & {\color{black}194} & 1000 & 1000 \\
    LDPS \citep{psld} & {\color{black}190}  & 1000& 1000 \\
    \midrule
    \rowcolor{gray!10}
    DPS ~\cite{dps}  & 180 & 1000 &1000 \\
    \rowcolor{gray!10}
    DMPS~\cite{meng2022diffusion} & 67 & 1000& 1000 \\    
    \bottomrule
  \end{tabu}%
  \hspace{5pt}\raisebox{-45pt}{\includegraphics[height=100pt]{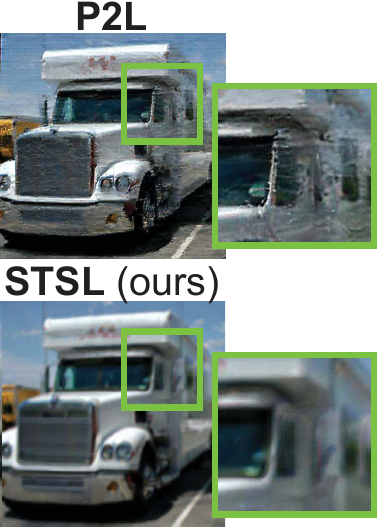}}
  \vspace{-1ex}
  \caption{
    {\bf(left)} Efficiency of LDM (top 4 rows) and PDM solvers (bottom 2 rows) on the super-resolution 8X task. {\bf(right)} Comparison of the image quality. The P2L image has LPIPS/SSIM of 0.51/74, and ours are 0.47/71.
  }
  \vspace{-0.2in}
  \label{tab:runtime-nfes}
\end{table}

\vspace{-0.5ex}
\subsection{Results on Image Inversion}
\label{sec:exps-inverse}
\vspace{-0.05in}
{\bf Inversion Quality:}
We evaluate our method against SoTA LDM-based solvers \citep{psld,p2l,he2023iterative} and PDM solvers \cite{dps,zhu2023denoising} 
on FFHQ and ImageNet datasets. 
Table \ref{tab:ffhq-imagenet} summarizes the results, highlighting STSL's ability in restoring the perceptual similarity of the original image, with improvement over SoTA P2L \cite{p2l} and PSLD \cite{psld}, particularly evident in LPIPS scores.
Table \ref{tab:ffhq-imagenet} also underscores the prevailing trend in LDM-based solvers \citep{psld,p2l} compared to PDM-based solvers \cite{dps,pGDM,ddrm}.
Notably, STSL surpasses P2L and PSLD with a 5\% absolute improvement in LPIPS on the demanding 8x super-resolution task.
Figure \ref{fig:exp:inversion} shows that STSL produces sharper and more detailed images without introducing artifacts or hallucination.

We adopt hyperparameters following the convention of inverse problem solvers \citep{dps,psld,p2l}, optimizing for LPIPS as it aligns with human perception. 
While maintaining competitive results on PSNR/SSIM, we notice our SSIM score could be less satisfactory in some cases. 
Our exploration reveals SSIM's inclination to label high-frequency artifacts as ``sharpness'' and its tendency to penalize blurriness more. In Table~\ref{tab:runtime-nfes} (right), the SSIM score of the P2L output is much better than ours regardless of the artifacts.
Consequently, the following discussion primarily emphasizes LPIPS.

{\bf Inversion Efficiency:}
Table \ref{tab:runtime-nfes} (left) compares solver efficiency amongst SoTA methods.
LDM-based solvers generate 512$\times$512 images whereas PDMs produce 256$\times$256. 
We downscale the LDM runtime by 4X for a fair comparison with PDMs. 
P2L \citep{p2l} runtime is estimated based on its pseudo-code due to the unavailability of source code.
Consequently, STSL achieves desired results in fewer diffusion steps ($T\!=\!50$) compared to PSLD \citep{psld} and P2L \citep{p2l} ($T\!=\!1000$).
This computational gain allows extra budget for local iterative gradient updates ($K=5$) using our surrogate loss.
In addition, STSL realizes a notable 4X improvement in the number of NFEs compared to PSLD \citep{psld}, and 8X over P2L \citep{p2l}.
Since NFE is the most expensive component in posterior sampling (\S\ref{sec:background}), less NFEs translate to practical advantages in runtime efficiency.

\begin{table*}[!t]
\centering
\setlength{\tabcolsep}{2pt}
\resizebox{\textwidth}{!}{%
\begin{tabular}{lccccccccccccccr}
\toprule
{} & \multicolumn{3}{c}{\textbf{Gaussian Deblur}} & \multicolumn{3}{c}{\textbf{Salt-pepper Denoise}} & \multicolumn{3}{c}{\textbf{SR ($\times 4$)}} & \multicolumn{3}{c}{\textbf{Random Inpaint}} & \multicolumn{3}{c}{\textbf{SR($\times 8$)}}
\\
\cmidrule(lr){2-4}
\cmidrule(lr){5-7}
\cmidrule(lr){8-10}
\cmidrule(lr){11-13}
\cmidrule(lr){14-16}
{\textbf{Method}} & {LPIPS$\downarrow$} & {PSNR$\uparrow$} & {SSIM$\uparrow$} & {LPIPS$\downarrow$} & {PSNR$\uparrow$} & {SSIM$\uparrow$} & {LPIPS$\downarrow$} & {PSNR$\uparrow$} & {SSIM$\uparrow$} & {LPIPS$\downarrow$} & {PSNR$\uparrow$} & {SSIM$\uparrow$}& {LPIPS$\downarrow$} & {PSNR$\uparrow$} & {SSIM$\uparrow$}\\
\midrule
\rowcolor{Orange!10}
STSL~(Ours) & \textbf{0.380} & \textbf{30.82} & \textbf{91.55} & \textbf{0.322} & \textbf{30.43} & \textbf{89.95} & \textbf{0.326} & \textbf{30.58} & \textbf{91.67} &\textbf{ 0.405 }& \textbf{30.13} & \textbf{88.24} & \textbf{0.398} & \textbf{30.32} & \textbf{88.12}\\
\rowcolor{Orange!10}
STSL-biased~(Ours) & \underline{0.456} & \underline{30.81} & \underline{90.70} & \underline{0.378} & \underline{30.29} & \underline{88.12} & \underline{0.406} & \underline{30.52} & \underline{89.91} & \underline{0.547} & \underline{29.64} & \underline{83.43} & \underline{0.444} & \underline{30.26} & \underline{85.18}\\
\cmidrule(l){1-13}
PSLD~\citep{psld} & 0.569 & 29.65 & 75.19 & 0.530 & 29.79 &78.74 & 0.531 & 29.84 & 80.22 & 0.693 & 28.86 & 67.25 & 0.538 & 29.71 & 77.71\\
\bottomrule
\end{tabular}
}
\vspace{-1ex}
\caption{
{\bf Quantitative results for bias analysis:}
All methods use 50 diffusion steps. 
STSL uses 5 stochastic averaging steps. 
STSL-biased uses a single step gradient update. The results are obtained with noise level $\sigma_\vy=0.05$. 
}
\vspace{-1ex}
\label{tab:ffhq-100}
\end{table*}

\vspace{-0.5ex}
\subsection{Ablation Study}
\vspace{-1ex}
\label{sec:ablation}

\noindent\textbf{Bias Analysis:}
STSL largely alleviates the bias existed in the first-order Tweedie estimator. 
We use stochastic averaging steps with the surrogate loss $\gL(\vy,Z_t)$ (\S\ref{sec:algo-inv}) and sample with an alternative reverse process (\ref{eq-reverse-posterior-refine}) initialized at $Z_0\sim p_T(Z_0|\enc(\mA^T\vy))$. By removing Hutchinson's trace estimator (setting $\eta=0$) and using a single step gradient update with $\gL(\vy,Z_t)$, we obtain the biased estimator STSL-biased. 
The distinction between STSL-biased and PSLD~\citep{psld} is that the former is initialized at $Z_0 \!\sim\! p_T(Z_0|\enc(\mA^T\vy))$ and the latter, $Z_0 \!\sim\! \pi_d$.
Based on 100 random images in FFHQ, Table~\ref{tab:ffhq-100} shows the benefits of using our {\em improved initialization} and {\em stochastic averaging steps} compared to prior methods~\citep{psld}. 
In summary, our new initialization enhances quality, and the stochastic steps aid faster convergence through local search.

\begin{table*}[!t]
\small
\raisebox{19pt}{\bf (a)}
\hspace{-13pt}\raisebox{-38pt}{\bf (b)}
\begin{minipage}[t]{0.25\textwidth}
  \begin{tabu} to \linewidth {X[l 0.3]*{3}{X[c 0.3]}}
    \toprule
    $N$ & LPIPS & PSNR & SSIM \\
    \midrule
    1 & 0.408 & 29.45 & 80.09\\
    \textbf{2} & \textbf{0.403} & 29.46 & 81.83\\
    3 & 0.405 & \textbf{29.51} & \textbf{81.89}\\
    \bottomrule
  \end{tabu}
  \hspace{0.01cm}
  \begin{tabu} to \linewidth  {X[l 0.3]*{3}{X[c 0.3]}}
    \toprule
    $\eta $ & LPIPS & PSNR & SSIM \\
    \midrule
    0 & 0.404 & 29.52 & 81.94\\
    0.02 & \textbf{0.388} & 29.52 & \textbf{82.70}\\
    0.05 & 0.395 & \textbf{29.54} & 82.45\\
    \bottomrule
  \end{tabu}
\end{minipage}
\raisebox{19pt}{\bf (c)}
\hspace{-13pt}\raisebox{-38pt}{\bf (d)}
\begin{minipage}[t]{0.7\textwidth}
  \begin{tabu} to 0.5\linewidth {X[l 0.3]*{5}{X[c 0.24]}}
    \toprule
    $\nu$ & 0.5 & 1 & 2 & 5 & 10\\
    \midrule
    LPIPS & 0.399 & 0.397 & \textbf{0.392} & 0.401 & 0.413\\
    PSNR & \textbf{29.44} & 29.41 & 29.39 & 29.31 & 29.19\\
    SSIM & 81.42 & 81.46 & \textbf{82.09} & 81.07 & 80.36\\
    \bottomrule
  \end{tabu}
  \raisebox{19pt}{\bf (e)}
  \begin{tabu} to \linewidth {lccccr}
    \toprule
    $K$ & LPIPS & PSNR & SSIM & NFE \\
    \midrule
    2 & 0.432 & \textbf{29.51} & 80.13 & 20 \\
    5 & \textbf{0.408} & 29.45 & \textbf{90.09} & 50 \\
    10& 0.424 & 29.33 & 77.42 & 100 \\
    \bottomrule
  \end{tabu} \\
  \hspace{0.01cm}
  \begin{tabu} to 0.33\linewidth {X[l 0.3]*{3}{X[c 0.24]}}
    \toprule
    Steps & 5 & 10 & 50  \\
    \midrule
    LPIPS & 0.564 & 0.511& \textbf{0.408}  \\
    PSNR & 28.56 & 28.87& \textbf{29.45}  \\
    SSIM & 66.91 & 73.57& \textbf{80.09}  \\
    \bottomrule
  \end{tabu}
  \raisebox{19pt}{\bf (f)}
  \begin{tabu} to 0.614\linewidth {X[l 2]@{\;}X[c 2.4]@{}X[c 2.4]@{}X[c 1.3]@{}X[c 2.4]}
    \toprule
    Method & NTI (clean) & NTI-CAT & NTI & STSL-CAT \\
    \midrule
    LPIPS & 0.434 & \textbf{0.425} & 0.638 & \textbf{0.552} \\
    PSNR & 29.06 & \textbf{29.11} & 28.74 & \textbf{28.81} \\
    CLIP acc. & \textbf{96.00} & \textbf{96.00} & 70.00 & \textbf{93.00}\\
    \bottomrule
  \end{tabu}
\end{minipage}
\vspace{-1.5ex}
\caption{
{\bf Ablation studies}:
(a) the number of samples $\veps \!\sim\! \pi_d$ used in stochastic averaging (\S\ref{sec:algo-inv}) at each diffusion step, 
(b) coefficient of the second-order approximation term in Eq.~\eqref{eq-surrogate-loss} with a single $\veps$,
(c) coefficient of the contrastive loss, 
(d) the number of DDIM steps,
(e) the number of the stochastic averaging steps, and 
(f) image editing results compared with NTI \cite{nti}, where the first two columns ``NTI (clean)" and ``NTI-CAT" are with clean images to show the effectiveness of CAT, while the last two column are on corrupted images with SRx8.
} 
\vspace{-1ex}

\label{tab:ablation}
\vspace{-0.1in}
\end{table*}




\noindent\textbf{Component Analysis:}
We study the significance of each component in STSL using 50 random samples from the COCO \citep{coco} dataset. 
Hyperparameters derived from this study also generalize to FFHQ-1K \cite{ffhq} and ImageNet-1K \cite{imagenet} datasets, showing the robustness of our method in handling unseen domains.
Throughout this analysis, we use salt-pepper noise as corruption in all the experiments.

Table~\ref{tab:ablation}(a) shows that employing a 2-sample average yields a more accurate estimate of the expectation in our surrogate loss $\gL(\vy,Z_t)$. 
With 5 stochastic averaging steps, we have $N=10$ Gaussian samples ($\veps$) in total per diffusion time step.
More samples lead to longer running time and higher memory demand with marginal benefits in LPIPS, but sharper image quality as evident from PSNR/SSIM.

Table~\ref{tab:ablation}(b) shows the Hutchinson's trace estimator, denoted as $\eta$, plays a crucial role. 
Optimal results were achieved at $\eta=0.02$. 
A value too small fails to mitigate the bias discussed in \S\ref{sec:algo-inv}, while too large values deviate the forward process, resulting in a different image.

Table \ref{tab:ablation}(c) demonstrates the effectiveness of the contrastive loss controlled by $\nu$ introduced in \S\ref{sec:algo-inv}.
As discussed in \S\ref{sec:algo-inv} and \S\ref{sec:algo-edit}, the contrastive loss helps preserve the \textit{content} of the source image through cross attention tuning (CAT) for both inverse problems and image editing. 

Tables \ref{tab:ablation}(d) and (e) study the impact of the diffusion steps and stochastic averaging steps. 
While the diffusion steps seem to be the larger the better, we observe 50 DDIM steps suffice for achieving satisfactory results. 
For the latter, too many stochastic averaging steps result in deviation from the small neighborhood around the forward latent. 
Thus, we use a modest number of stochastic averaging steps (\ie, 5) of proximal gradient update, as detailed in \S\ref{sec:algo-inv}.

\subsection{Results on Image Editing}
\vspace{-0.5ex}
\label{sec:exps-editing}

{\bf Qualitative Study:}
STSL seamlessly extends to editing corrupted image.
As shown in Figure \ref{fig:exp:edit}, both NTI \citep{nti} and commercial software exhibit lower editing quality for real images with or without corruption. 
NTI \citep{nti} struggles to generate quality images with corruptions. 
The commercial software demands user-selected regions for editing.
In contrast, STSL-CAT accurately localizes the edits without user intervention, and preserves the integrity of the entire image.
For example, in NTI \citep{nti}, replacing the car with a bicycle affects the surrounding house, and transforming the house to a supermarket changes the SUV into a sedan.

{\bf Quantitative Study:}
Table~\ref{tab:ablation}(f) shows a quantitative analysis of 100 randomly selected dog images from ImageNet for ``a dog'' to ``a cat'' editing. 
NTI-CAT incorporates NTI \citep{nti} with our Cross-Attention-Tuning (\S\ref{sec:algo-edit}) in the reverse process, and shows consistently improvement over NTI on the clean image editing (the first two columns). 
The CLIP accuracy \citep{hessel2021clipscore} remains similar, because it doesn't measure content preservation but only the matching between the output and the target prompt.
In corrupted image editing (the last two columns), our end-to-end STSL-CAT pipeline demonstrates effectiveness by surpassing NTI \citep{nti} with a notable relative improvement of  32\% in CLIP accuracy.

\vspace{-1ex}
\section{Conclusion}
\vspace{-1ex}
\label{sec:conc}

We have introduced STSL, a novel posterior sampler that matches the efficiency of the first-order Tweedie with a tractable second-order approximation in a new reverse process.
Our theoretical results reveal that the second-order approximation can be lower-bounded by our surrogate loss that requires only an estimate of the trace of Hessian, leading to 4X and 8X reduction in neural function evaluations compared to SoTA solvers PSLD \cite{psld} and P2L \cite{p2l}, respectively, while enhancing sampling quality on various inversion tasks.
STSL seamlessly extends to text-guided image editing, surpassing a leading solution NTI \cite{nti} especially in handling corrupted images.
To our best knowledge, this marks the first {\em efficient} second-order approximation in solving inverse problems using latent diffusion and image editing with corruption.

\noindent\textbf{Limitation:} The proposed inverse problem solver uses $\mA^T$ from DPS~\citep{dps}, which is set to identity for some tasks. It might be better to use Jax implementation of $A^T$ for improved performance as in P2L~\citep{p2l}.

\noindent\textbf{Future work:} Our approach does not tune the prompt used in the generative foundation model. Integrating prompt-tuning~\cite{p2l} into our pipeline might prove beneficial.

\noindent\textbf{Reproducibility:} The pseudo-code of STSL for inverse is given in \textbf{Algorithm~\ref{alg:stsl}} and editing in \textbf{Algorithm~\ref{alg:stsl-edit}}. All the hyper-parameter details are provided in \S\ref{sec:exps} and \S\ref{sec:impl-details}.

\section*{Acknowledgements}

 This research has been partially supported by NSF Grant 2019844, Google Research, and the UT Austin Machine Learning Lab (MLL). 
 Litu Rout has been supported by the Ju-Nam and Pearl Chew Endowed Presidential Fellowship in Engineering and the George J. Heuer, Jr. Ph.D. Endowed Graduate Fellowship.

{
    \small
    \bibliographystyle{ieeenat_fullname}
    \bibliography{main}
}


\clearpage
\appendix
\onecolumn

\section*{Appendix}
This section includes more results and details that did not fit into the main paper due to space limitation.
Particularly, we offer expanded theoretical analysis in \S\ref{sec:app-theory} and implementation details in \S\ref{sec:addn-exps}, along with other supportive analysis. These sections provide a deeper understanding and comprehensive context to the research presented in the main body of the paper.


\section{Theoretical Analysis}
\label{sec:app-theory}

\subsection{Posterior mean and covariance using Tweedie's formula}
\label{sec:prf-tweedie}
\begin{proposition}[\citep{robbins1956empirical,efron2011tweedie}]
    \label{prop-tweedie-long}
    Given $\vx_t = \sqrt{\bar{\alpha}_t}\vx_0 + \sqrt{1-\bar{\alpha}_t}\mathbf{\veps}$ and $\veps \sim \gN\left(  0, I\right)$, denote by $\bar{X}_0=\E_{X_0 \sim p_{t}\left( X_0|X_t=\vx_t\right)}[X_0]$ the posterior mean of $p_t(X_0|X_t=\vx_t)$. Then, for the variance preserving SDE or DDPM sampling, $p_t(X_0|X_t=\vx_t)$ has mean
    $$\E_{X_0 \sim p_t(X_0|X_t=\vx_t)}\left[X_0\right] = \frac{\vx_t}{\sqrt{\bar{\alpha}_t}}+\frac{(1-\bar{\alpha}_t)}{\sqrt{\bar{\alpha}_t}}\nabla_{\vx_t} \log p_t(X_t=\vx_t)$$
    and covariance
    $$\E_{X_0 \sim p_t(X_0|X_t=\vx_t)}\left[\left(X_0-\bar{X}_0\right)\left(X_0-\bar{X}_0\right)^T\right] = \frac{1-\bar{\alpha}_t}{\bar{\alpha}_t}\left(  \mI + (1-\bar{\alpha}_t)\nabla_{\vx_t}^2 \log p_t(X_t=\vx_t) 
    \right).$$
\end{proposition}

\begin{proof}
    Given $\vx_t = \vmu + \sigma \veps$, where $\veps \sim \gN\left(\vzero,\mI\right)$, we know that $\vx_t \sim \gN\left( \vmu, \sigma^2\mI\right)$. From [\citealp[Section 2]{efron2011tweedie}], we have 
    \begin{align*}
        \E\left[\vmu|\vx_t \right] 
        & 
        = \vx_t + \sigma^2 \nabla_{\vx_t} \log p_t(\vx_t),
        \\
        \mathbb{V}\left[\vmu | \vx_t \right]
        & = \sigma^2 \left(1+\sigma^2 \right) \nabla_{\vx_t} \log p_t(\vx_t),
    \end{align*}
    where $\E\left[\vmu|\vx_t \right]$  and $\mathbb{V}\left[\vmu | \vx_t \right]$ denote the conditional mean and the conditional variance, respectively. Since $\vx_t = \sqrt{\bar{\alpha}_t}\vx_0 + \sqrt{1-\bar{\alpha}_t}\mathbf{\veps}$ in our case, we get
    \begin{align*}
        \E\left[\sqrt{\bar{\alpha}_t}\vx_0|\vx_t \right] 
        & 
        = \sqrt{\bar{\alpha}_t}\E\left[\vx_0|\vx_t \right] 
        = \vx_t + \left(1-\bar{\alpha}_t \right) \nabla_{\vx_t} \log p_t(\vx_t),
        \\
        \mathbb{V}\left[\sqrt{\bar{\alpha}_t}\vx_0 | \vx_t \right]
        & 
        = \bar{\alpha}_t \mathbb{V}\left[\vx_0 | \vx_t \right]
        = \left(1-\bar{\alpha}_t \right) \left(1+\left(1-\bar{\alpha}_t \right)\right) \nabla_{\vx_t} \log p_t(\vx_t),
    \end{align*}
    which upon rearrangement yields the following:
    \begin{align*}
    \E_{X_0 \sim p_t(X_0|X_t=\vx_t)}\left[X_0\right] 
    & 
    =
    \frac{\vx_t}{\sqrt{\bar{\alpha}_t}}+\frac{(1-\bar{\alpha}_t)}{\sqrt{\bar{\alpha}_t}}\nabla_{\vx_t} \log p_t(X_t=\vx_t)\\
    \E_{X_0 \sim p_t(X_0|X_t=\vx_t)}\left[\left(X_0-\bar{X}_0\right)\left(X_0-\bar{X}_0\right)^T\right] 
    &
    =
    \frac{1-\bar{\alpha}_t}{\bar{\alpha}_t}\left(  \mI + (1-\bar{\alpha}_t)\nabla_{\vx_t}^2 \log p_t(X_t=\vx_t) 
    \right).
    \end{align*}
    This completes the proof of the statement.
\end{proof}

\subsection{First-order Tweedie sampler}
\label{sec:thm-dps}
\begin{theorem}(First-order Tweedie Estimator~\citep{dps}).
    \label{thm-dps}
    Given measurements $\vy = \gA (\vz_T) + \vn$, $\vn\sim \gN\left( \vzero,\sigma^2_\vy \mI\right)$ and the first-order approximation $p_{T-t}(\vy|Z_t)\approx p_{T-t}(\vy|\bar{Z}_T)$, define the Jensen's gap as: 
    $$\gJ \coloneqq \left| \E_{Z_T\sim p_{T-t}\left(Z_T|Z_t\right)}[p_{T-t}(\vy|Z_t)]-p_{T-t}(\vy|\bar{Z}_T) \right|,$$
    where $\bar{Z}_T\coloneqq\E_{Z_T \sim p_{T-t}(Z_T|Z_t)}\left[ Z_T\right]$. Then, the error due to first-order approximation is upper bounded by 
    \begin{align*}
         \gJ \leq \frac{d}{\sqrt{2\pi \sigma^2_\vy}} \exp\left(-\frac{1}{2\sigma^2_\vy}\right)\|\nabla_{z} \gA(\vz)\|m_1,
    \end{align*}
    where $\|\nabla_{\vz} \gA(\vz)\|\coloneqq \max_{\vz}\| \gA(\vz)\|$ and $m_1\coloneqq \int \|Z_T - \bar{Z}_T\|~p_{T-t}(Z_T|Z_t)dZ_T$.
\end{theorem}

Since $\|\nabla_{\vz} \gA(\vz)\|$ and $m_1$ are finite for most inverse problems,
the Jensen's gap goes to zero as $\sigma_\vy \rightarrow \infty$, leading to less approximation error in (\ref{eq-mean-approx}). This setting is of less practical significance because as $\sigma_\vy\rightarrow\infty$, the measurements $\vy=\gA(\vz_T) + \sigma_\vy \veps, \veps\sim \gN\left( \vzero, \mI\right)$ provide no meaningful information about $\vz_T$. Thus, sampling from the posterior $p_{0}(Z_T|\vy)=p_{0}(X_0|\vy)$ is as good as sampling from the prior $p_0(X_0)$. On the other hand, when $\sigma_\vy \rightarrow 0$, the problem is reduced to a noiseless setting which is relatively easier to deal with. In practically relevant settings where $\sigma_y$ is non-zero and finite, the Jensen's gap could be arbitrarily large.
This leads to a bias in reconstruction and sub-optimal performance in various tasks as we show in \S\ref{sec:exps}.

\subsection{Second-order Tweedie sampler from surrogate loss}
\label{sec:prf-tsl}

\begin{theorem}[Tweedie Sampler from Surrogate Loss]
    \label{thm:unbiased-tweedie-long}
    Suppose \textbf{Assumption~\ref{assm-meas-exp}} and \textbf{Assumption~\ref{assm-meas-func}} hold. Let $\hat{\gL}(\vy,Z_t)$ denote the function:
    $$\hat{\gL}(\vy,Z_t)\coloneqq 
    \log \left( p_{T-t}(\vy|\bar{Z}_T) \right)
    +
    \log \left( 1-
    \frac{1-\bar{\alpha}_t}{\bar{\alpha}_t} md
    -
    (1-\bar{\alpha}_t) m~Trace\left(\nabla^2 \log p_{T-t}(Z_t) \right)
    \right).$$
    For $\lambda=\gO(\frac{1}{\sigma_\vy^2})$ and $\gamma = \gO(\frac{\eta}{d})$, the following holds:
    $\hat{\gL}(\vy,Z_t) \leq \log p_{T-t}\left(\vy|Z_t\right).$
    Further, the gradient of $\hat{\gL}(\vy,Z_t)$ is given by:
    $$\nabla \hat{\gL}(\vy,Z_t)
    =
    -\frac{1}{2\sigma_\vy^2}  \nabla \lVert\vy-\mA\bar{Z}_T \rVert^2
    - \frac{(1-\bar{\alpha}_t) m }{\left( 1-
    \frac{1-\bar{\alpha}_t}{\bar{\alpha}_t} md
    -
    (1-\bar{\alpha}_t) m~Trace\left(\nabla^2 \log p_{T-t}(Z_t) \right)
    \right)}
    \nabla Trace\left(\nabla^2 \log p_{T-t}(Z_t) \right).$$
\end{theorem}

\begin{proof}
We want to compute the following:
    \begin{align}
    \nonumber
    \log p_{T-t}(\vy|Z_t) &= \log \int p_{T-t}(\vy|Z_t,Z_T)  p_{T-t}(Z_T|Z_t)dZ_T\\
    \nonumber
    &\stackrel{(i)}{=}  \log \int p_{T-t}(\vy|Z_T)  p_{T-t}(Z_T|Z_t)dZ_T\\
    &=  \log \E_{Z_T \sim p_{T-t}(Z_T|Z_t)}\left[p_{T-t}(\vy|Z_T)\right]
    \label{eq-tr-1}
\end{align}
where (i) is because $\vy$ is independent of $Z_t$ given $Z_T$. Denote by $\bar{Z}_T = \E_{Z_T \sim p_{T-t}(Z_T|Z_t)}\left[Z_T\right]$. Now, using Taylor series expansion at $\bar{Z}_T$, for some $\tilde{Z}_T \in \gB_r\left(\bar{Z}_T\right)\coloneqq \{Z \in \R^d| \lVert Z-\bar{Z}_T \rVert \leq r\}, r = \lVert Z_T-\bar{Z}_T\rVert$, we get 
\begin{align}
    \nonumber
    &
    \log \E_{Z_T \sim p_{T-t}(Z_T|Z_t)}\left[p_{T-t}(\vy|Z_T)\right] \\
    \nonumber
    &
    = \log \E_{Z_T \sim p_{T-t}(Z_T|Z_t)}\left[p_{T-t}(\vy|\bar{Z}_T) + \langle \nabla p_{T-t}(\vy|Z_t)\mid_{\bar{Z}_T}, Z_T - \bar{Z}_T  \rangle + \frac{1}{2}(Z_T - \bar{Z}_T )^T\nabla^2 p_{T-t}(\vy|\tilde{Z}_T) (Z_T - \bar{Z}_T ) \right]
    \\
    \nonumber
    &
    = 
    \log \left( p_{T-t}(\vy|\bar{Z}_T) + \frac{1}{2}\E_{Z_T \sim p_{T-t}(Z_T|Z_t)}\left[ (Z_T - \bar{Z}_T )^T\nabla^2 p_{T-t}(\vy|\tilde{Z}_T) (Z_T - \bar{Z}_T ) \right] \right),
\end{align}
where the last step follows from linearity of expectation and the fact that $\langle \nabla p_{T-t}(\vy|Z_t)\mid_{\bar{Z}_T}, \E_{Z_T \sim p_{T-t}(Z_T|Z_t)}\left[Z_T\right] - \bar{Z}_T  \rangle =0$. Since $\log (a+b) = \log(a) + \log (1+ b/a)$ for $a>0$ and $p_{T-t}(\vy|\bar{Z}_T)>0$ due to \textbf{Assumption~\ref{assm-meas-exp}}, the above expression simplifies to
\begin{align}
    \nonumber
    &
    \log \E_{Z_T \sim p_{T-t}(Z_T|Z_t)}\left[p_{T-t}(\vy|Z_T)\right] 
    \\
    \nonumber
    &
    = 
    \log \left( p_{T-t}(\vy|\bar{Z}_T) \right)
    +
    \log \left(1+ \frac{ \E_{Z_T \sim p_{T-t}(Z_T|Z_t)}\left[ (Z_T - \bar{Z}_T )^T\nabla^2 p_{T-t}(\vy|\tilde{Z}_T) (Z_T - \bar{Z}_T ) \right]}{2 p_{T-t}(\vy|\bar{Z}_T) } \right) 
    \\
    \nonumber
    &
    =\log \left( p_{T-t}(\vy|\bar{Z}_T) \right)
    +
    \log \left( 1+ \E_{Z_T \sim p_{T-t}(Z_T|Z_t)}\left[ (Z_T - \bar{Z}_T )^T\left( \frac{\nabla^2 p_{T-t}(\vy|\tilde{Z}_T)}{2p_{T-t}(\vy|\bar{Z}_T)}\right) (Z_T - \bar{Z}_T ) \right] \right)
    \\
    \nonumber
    & 
    \geq \log \left( p_{T-t}(\vy|\bar{Z}_T) \right)
    +
    \log \left( 1 - m ~\E_{Z_T \sim p_{T-t}(Z_T|Z_t)}\left[ (Z_T - \bar{Z}_T )^T (Z_T - \bar{Z}_T ) \right] \right)
    \\
    \nonumber
    &
    = \log \left( p_{T-t}(\vy|\bar{Z}_T) \right)
    + 
    \log  \left(1-m~Trace  \left( \E_{Z_T \sim p_{T-t}(Z_T|Z_t)}\left[\left(Z_T-\bar{Z}_T\right)\left(Z_T-\bar{Z}_T\right)^T\right]  \right) \right)
    \\
    \nonumber
    &
    = \log \left( p_{T-t}(\vy|\bar{Z}_T) \right)
    +
    \log \left(1-m~Trace  \left( \frac{1-\bar{\alpha}_t}{\bar{\alpha}_t}\left(  I + (1-\bar{\alpha}_t)\nabla^2 \log p_{T-t}(Z_t) 
    \right)  \right) \right)
    \\
    \nonumber
    &
    = \log \left( p_{T-t}(\vy|\bar{Z}_T) \right)
    +
    \log \left( 1-
    \frac{1-\bar{\alpha}_t}{\bar{\alpha}_t} md
    -
    (1-\bar{\alpha}_t) m~Trace\left(\nabla^2 \log p_{T-t}(Z_t) \right)
    \right) \coloneqq \hat{\gL}(\vy,Z_t)
\end{align}
This completes the proof of the first part, $\hat{\gL}(\vy,Z_t) \leq \log p_{T-t}\left(\vy|Z_t\right)$. 

Next, the gradient of the lower bound with respect to $Z_t$ becomes:
\begin{align}
    \nonumber
    &
    \nabla \hat{\gL}(\vy,Z_t)
    \\
    \nonumber
    &
    = -\frac{1}{2\sigma_\vy^2}  \nabla \lVert\vy-\mA\bar{Z}_T \rVert^2
    +
    \nabla \log \left( 1-
    \frac{1-\bar{\alpha}_t}{\bar{\alpha}_t} md
    -
    (1-\bar{\alpha}_t) m~Trace\left(\nabla^2 \log p_{T-t}(Z_t) \right)
    \right)
    \\
    \nonumber
    &
    = -\frac{1}{2\sigma_\vy^2}  \nabla \lVert\vy-\mA\bar{Z}_T \rVert^2
    - \frac{(1-\bar{\alpha}_t) m }{\left( 1-
    \frac{1-\bar{\alpha}_t}{\bar{\alpha}_t} md
    -
    (1-\bar{\alpha}_t) m~Trace\left(\nabla^2 \log p_{T-t}(Z_t) \right)
    \right)}\nabla Trace\left(\nabla^2 \log p_{T-t}(Z_t) \right),
\end{align}
where the last step follows from $ \nabla\left( 1-\frac{1-\bar{\alpha}_t}{\bar{\alpha}_t} md\right)=0$. 
\end{proof}
\noindent \textbf{Implication:} From the above result, we have
\begin{align*}
    \nabla \hat{\gL}(\vy,Z_t)
    &
    \simeq
    -\lambda  \nabla \lVert\vy-\mA\bar{Z}_T \rVert^2
    - \gamma 
    \nabla  \left(
    Trace\left(\nabla^2 \log p_{T-t}(Z_t) \right)
    \right),
\end{align*}
where $\lambda = \gO\left(\frac{1}{\sigma_\vy^2}\right)$ and $\gamma = \gO\left(\frac{\eta}{d}\right)$ are hyper-parameters to be tuned in practice. 

\noindent\textbf{Connection with the surrogate loss:} The gradient of the lower bound $\hat{\gL}(\vy,Z_t)$ is equal to the negative gradient of the surrogate loss function $\gL(\vy,Z_t)$ introduced in \S\ref{sec:algo-inv} and \S\ref{sec:theory}, i.e., $\nabla \hat{\gL}(\vy,Z_t) \simeq -\nabla \gL(\vy,Z_t)$, when the constants $\lambda$ and $\gamma$ are chosen appropriately.
More precisely, as given in the statement of the \textbf{Theorem~\ref{thm:unbiased-tweedie-long}}, these gradients are equal when $\lambda = \frac{-1}{2\sigma_\vy^2}$ and $\gamma = \frac{-(1-\bar{\alpha}_t) m }{\left( 1-
    \frac{1-\bar{\alpha}_t}{\bar{\alpha}_t} md
    -
    (1-\bar{\alpha}_t) m~Trace\left(\nabla^2 \log p_{T-t}(Z_t) \right)
    \right)}$.
In our implementation, we use $\nabla \gL(\vy,Z_t)$ that results in proximal gradient \textit{descent} in \textbf{Algorithm~\ref{alg:stsl}}.

\subsection{Computation using Hutchinson's Trace Estimator}
\label{sec:prf-hutchinson}

    Given $\veps\sim \gN\left( \vzero, \mI \right)$, the trace of the Hessian can be efficiently computed as:
    $$
    \E\left[ \veps^T \left(\nabla \log p_{T-t}\left(Z_t  + \veps\right) - \nabla \log p_{T-t}\left(Z_t\right)\right)
        \right] - \gO(\lVert\veps\rVert^3)
    \simeq
    Trace\left(\nabla^2 \log p_{T-t}(Z_t) \right).$$

    \noindent To see this, for $\veps \sim \gN\left( \vzero, \mI \right)$, using Taylor series expansion of the score, we get
    \begin{align*}
        \nabla \log p_{T-t}\left(Z_t  + \veps\right)  
        \simeq
        \nabla \log p_{T-t}\left(Z_t\right)
        + \nabla^2 \log p_{T-t}(Z_t) \veps + \gO\left( \lVert\veps\rVert^2\right).
    \end{align*}
    Subtracting $\nabla \log p_{T-t}\left(Z_t\right)$ from both sides, and taking projection onto $\veps$, we have 
    \begin{align*}
        \veps^T \left(\nabla \log p_{T-t}\left(Z_t  + \veps\right) - \nabla \log p_{T-t}\left(Z_t\right)\right) 
        \simeq \veps^T \nabla^2 \log p_{T-t}(Z_t) \veps
        + \gO(\lVert\veps\rVert^3).
    \end{align*}
    The claim follows by taking the expectation of both sides and applying Hutchinson's trace estimator~\citep{hutchinson1989stochastic} as given below: 
    \begin{align}
        \nonumber
        \E\left[ \veps^T \left(\nabla \log p_{T-t}\left(Z_t  + \veps\right) - \nabla \log p_{T-t}\left(Z_t\right)\right)
        \right] - \gO(\lVert\veps\rVert^3)
        & 
        \simeq 
        \E_{\veps \sim \gN\left( 0, I \right)}\left[ \left(\veps^T\nabla^2 \log p_{T-t}(Z_t) \veps\right) \right] 
        \\
        \nonumber
        & 
        =
        Trace\left(\nabla^2 \log p_{T-t}(Z_t) \right)
    \end{align}

The last step above involves an approximation of a higher derivative through an expectation of random projections of perturbed function evaluations. This approach has been well studied in online learning settings and with formal guarantees (e.g., Lemma~2.1 in \cite{flkamc05}). In our case, the approximation additionally involves a ``centering'' with $\veps^T \nabla \log p_{T-t}\left(Z_t\right)$. While this terms is zero in expectation, it is useful to keep because as we discuss in Section~\ref{sec:algo-inv}, we are evaluating the expectation through stochastic averaging with finitely many steps. This centering decreases the magnitude of each step, thus resulting in variance improvement (and thus a less noisy approximation with a fewer number of steps).


\section{Additional Experimental Evaluation}
\label{sec:addn-exps}
\subsection{Implementation Details}
\label{sec:impl-details}
\textbf{Image Inversion:} We follow the same experimental setup as prior works~\citep{dps,psld}, and use the measurement operators provided in their original source code: DPS\footnote{
\url{https://github.com/DPS2022/diffusion-posterior-sampling}}  and PSLD\footnote{\url{https://github.com/LituRout/PSLD}}.
We employ a Gaussian blur kernel (size $61\times61$, $\sigma=3.0$) for \textit{Gaussian deblurring} and a motion blur kernel (size $61\times61$, intensity 0.5) for \textit{motion deblurring} tasks.
For \textit{super-resolution}, we use $4\times$ and $8\times$ downsampling as measurement operator. 
Additionally, we introduce 2\% salt and pepper noise for \textit{denoising} and 40\% drop rate for \textit{random inpainting} tasks. For \textit{free-form inpainting}, we adopt the 10\%-20\% damage range as utilized in prior works~\citep{saharia2022palette, p2l}.

Our refinement module in \textbf{Algorithm~\ref{alg:stsl}} uses the Adam optimizer, with an initial learning rate of $1e-2$ and decrementing by a factor of 0.998 per diffusion time step. This process optimizes the latents with stochastic averaging. Notably, STSL exhibits robustness across various tasks, showing minimal sensitivity to hyper-parameter changes. Therefore, we maintain consistent configurations for all tasks, where $N=2$, $\eta=0.02$, $\nu=2$ and $\lambda=1$.
We use $K=5$ and $T=50$ as defualt and conduct extensive ablation studies for free-form image inpainting task in \S\ref{sec:more-quant-results}.
Following the experimental setting of P2L\citep{p2l}, we add independent and identically distributed Gaussian noise $\gN\left( \vzero,0.01^2\right)$ to each pixel. 

\noindent\textbf{Image Editing:}
In image editing, we use a single stochastic averaging step $K=1$ since the latents have been refined during proximal gradient updates. We use $\nu=0.02$ for the contrastive loss without normalization by the data dimension $d$, $\lambda=1$ for the measurement loss and the same coefficient for Hutchinson’s trace estimator $\eta=0.02$ as in inversion. More details are elaborated in \textbf{Algorithm~\ref{alg:stsl-edit}}. For the qualitative demonstration, we compare with NTI\footnote{\url{https://github.com/google/prompt-to-prompt/}} and a commercial platform that is publicly available. We conduct the experiments using the latest version of the commercial software by November 2023.

\begin{algorithm}[!t]
    \caption{Second-order Tweedie sampler from Surrogate Loss (STSL) for image inversion and editing task}
    \label{alg:stsl-edit}
         \KwIn{
         diffusion time steps $T$,
         observed $\vy$,
         measurement operator $\mA$,
         likelihood strength $\lambda$,
         stochastic averaging steps $K$,
         second-order correction stepsize $\eta$,
         encoder $\enc$,
         decoder $\dec$,    
         learned score  $\bm{s}_\theta
         $,\hspace{5cm}
         target text $``prompt"$,
         text encoder $\Phi$
         }
          Initialization: $\overrightarrow{Z}_0 = \enc(\mA^T\vy)$ \hfill $\triangleright$ forward process\\
          \For{$t=0$ \KwTo $T-1$}{
          $\overrightarrow{Z}_{t+1}
          \gets 
          \sqrt{\alpha_{t}}\overrightarrow{Z}_{t}
          -\sqrt{(1-\alpha_t)} \sqrt{(1-\bar{\alpha}_t)} s_\vtheta(\overrightarrow{Z}_{t}, t) $\\
          }
          Initialization: $Z_0 = \overrightarrow{Z}_T$ \hfill $\triangleright$ reverse process for image inversion\\
          \For{$t=0$ \KwTo $T-1$}{
            \For{$k=0$ \KwTo $K$}{
                $\veps \sim \gN(\bm{0}, \mI)$ \hfill $\triangleright$ stochastic averaging\\
                $\bar{Z}_T
                \gets 
                (Z_t + (1 - \bar\alpha_{T-t}) \bm{s}_\vtheta(Z_t, T-t))/ \sqrt{\bar\alpha_{T-t}}
                $\\
                $Z_{t} \gets Z_{t} - \lambda \nabla \lVert \vy - \mA\dec(\bar{Z}_T)\rVert
                - 
                (\eta/d)
                  \nabla \left(\veps^T \bm{s}_\vtheta\left(Z_t+ \veps,T-t\right)
                  - \veps^T\bm{s}_\vtheta\left(Z_t,T-t \right)
                  \right)$\\
            }
            $\bar{Z}_T
            \gets 
            (Z_t + (1 - \bar\alpha_{T-t}) \bm{s}_\vtheta(Z_t, T-t))/ \sqrt{\bar\alpha_{T-t}}
            $\\
            $Z_{t+1} 
            \gets
            \frac{\sqrt{\alpha_{T-t}}(1-\bar{\alpha}_{T-t-1})}{1 - \bar{\alpha}_{T-t}}Z_t
            + \frac{\sqrt{\bar{\alpha}_{T-t-1}}(1-\alpha_{T-t})}{1 -\bar{\alpha}_{T-t}}\bar{Z}_T 
            $\\
            }
            Initialization: $Z_0 = \overrightarrow{Z}_T$
            \hfill $\triangleright$ reverse process for image editing\\
            \For{$t=0$ \KwTo $T-1$}{
            $\bar{Z}_T
            \gets 
            (Z_t + (1 - \bar\alpha_{T-t}) \bm{s}_\vtheta(Z_t, T-t, \varphi_t))/ \sqrt{\bar\alpha_{T-t}}
            $\\
            $f\left(Z_t,T-t,\varphi_t \right) \!=\! \sqrt{\bar{\alpha}_{T-t-1}}\bar{Z}_T \!+\! \sqrt{1-\bar{\alpha}_{T-t-1}}\sqrt{1-\bar{\alpha}_{T-t}}\bm{s}_\vtheta(Z_t,T \!-\! t,\varphi_t)$\\
            $\hat{\varphi}_{t} = \arg \min_{\varphi_t} \lVert Z_{t+1} - f\left(Z_t,T-t,\varphi_t \right) \rVert_2^2$ \hfill $\triangleright$ Null-optimization\\
            $\hat{Z}_{t+1} \gets CAC(Z_t,T-t,\hat{\varphi}_t,\Phi\{ ``prompt"\})$ \hfill $\triangleright$ Cross-Attention-Control (CAC)~\citep{p2p}\\
            $\veps \sim \gN(\bm{0}, \mI)$\\
                $\bar{Z}_T
                \gets 
                (Z_t + (1 - \bar\alpha_{T-t}) \bm{s}_\vtheta(Z_t, T-t, \Phi\{ ``prompt"\}))/ \sqrt{\bar\alpha_{T-t}}
                $\\
                $Z_{t+1} \gets \hat{Z}_{t+1} - \lambda \nabla \lVert \vy - \mA\dec(\bar{Z}_T)\rVert
                - 
                \frac{\eta}{d}
                  \nabla \veps^T \left( \bm{s}_\vtheta\left(Z_t+ \veps,T-t,\Phi\{ ``prompt"\}\right)
                  - \bm{s}_\vtheta\left(Z_t,T-t,\Phi\{ ``prompt"\} \right)
                  \right)$\\
            }
            \Return $\dec(Z_T)$
\end{algorithm}

\subsection{Computational Complexity}
\label{sec:complexity}
Table~\ref{tab:runtime-nfes} provides a comparative analysis of the runtime performance across various state-of-the-art methods. NFEs are computed based on the required reverse and optimization steps. For instance, P2L~\citep{p2l} demands 1000 reverse steps, accompanied by at least one prompt tuning step per iteration, accumulating in a total of 2000 NFEs. The best results of P2L~\cite{p2l} are obtained with around 5000 NFEs, which amounts to \textbf{30 mins} of runtime per image. Other baseline methods require 1000 reverse steps. The best results of PSLD/GML-DPS~\citep{psld} are obtained with 1000 NFEs, which amounts to \textbf{12 mins} of runtime per image. Our STSL framework demonstrates \textit{efficiency} by employing only 50 DDIM steps coupled with 5 stochastic averaging steps, resulting in a considerably lower count of 250 NFEs. This translates into significantly lower runtime of \textbf{under 3 min} with a considerable gain in performance. Note that the runtime of PDM-solvers is lower because the underlying generative model is smaller compared to large-scale foundation models, such as Stable Diffusion. Despite smaller runtime, PDM-solvers are subpar SoTA solvers~\citep{psld,p2l} leveraging these foundation models.

\begin{table*}[!thb]
\centering
\small
\begin{tabular}{lcccccr}
\toprule
{\textbf{Method}} & {LPIPS$\downarrow$} & {PSNR$\uparrow$} & {SSIM$\uparrow$} & $K$ & $T$ & NFEs\\
\midrule
STSL-I~(Ours) & 0.279 & 30.61 & 81.53 & 5&50&250\\
STSL-III~(Ours) & 0.282  & 30.79 & 82.55 & 5 & 200 & 1000 \\
\midrule
\rowcolor{Orange!10}
STSL-II~(Ours) & 0.386 & 29.65 & 77.16 & 5&50&250\\
\rowcolor{Orange!10}
STSL-IV~(Ours) & \underline{0.311} & 30.29& 81.74 & 5 & 200 & 1000\\
\rowcolor{Orange!10}
STSL-V~(Ours) & \textbf{0.291} & 30.65 & 82.48 & 2 & 1000 & 2000\\
\cmidrule(l){1-7}
P2L~\citep{p2l} & 0.321 & 31.29 & \textbf{85.16} & 2 & 1000 & 2000 \\
PSLD~\citep{psld} & 0.344 & \textbf{31.54} & 84.20 & 1 & 1000 & 1000\\
 GML-DPS~\citep{psld} &0.364 & \underline{31.49} & 84.00 & 1 & 1000 & 1000\\
LDPS~\citep{psld} & 0.379 & 31.34& 84.45 & 1 & 1000& 1000\\
LDIR~\citep{he2023iterative} & 0.386 & 31.24& \underline{84.87} & 1 & 1000& 1000\\
\cmidrule(l){1-7}
\rowcolor{gray!10}
DPS~\citep{dps} & 0.368 & 28.96 & 69.89 & 1 & 1000& 1000\\
\bottomrule
\end{tabular}

\caption{
\textbf{Quantitative results of the free-form inpainting task on ImageNet-1K}. 
 STSL-I/III are initialized from the forward latent $Z_0\sim p_T(Z_0|\enc(\mA^T\vy))$ while all the other methods are initialized with Gaussian noise $Z_0\sim \pi_d$.
As discussed in \S\ref{sec:more-quant-results}, STSL-I/III sometimes leaves small missing areas as shown in Figure~\ref{fig:appendix:fip-fail} even though it better reconstructs unmasked regions of the image.
To make a fair comparison, we only consider the methods using the same initialization from the Gaussian noise that successfully inpaint all the missing regions.
In this setting, STSL-IV and STSL-V still outperform SoTA solver PSLD~\citep{psld} and P2L~\citep{p2l} using the same number of NFEs: 1000 and 2000, respectively.
}
\label{tab:appendix}
\end{table*}

\begin{figure*}
    \centering
    \includegraphics[width=\linewidth]{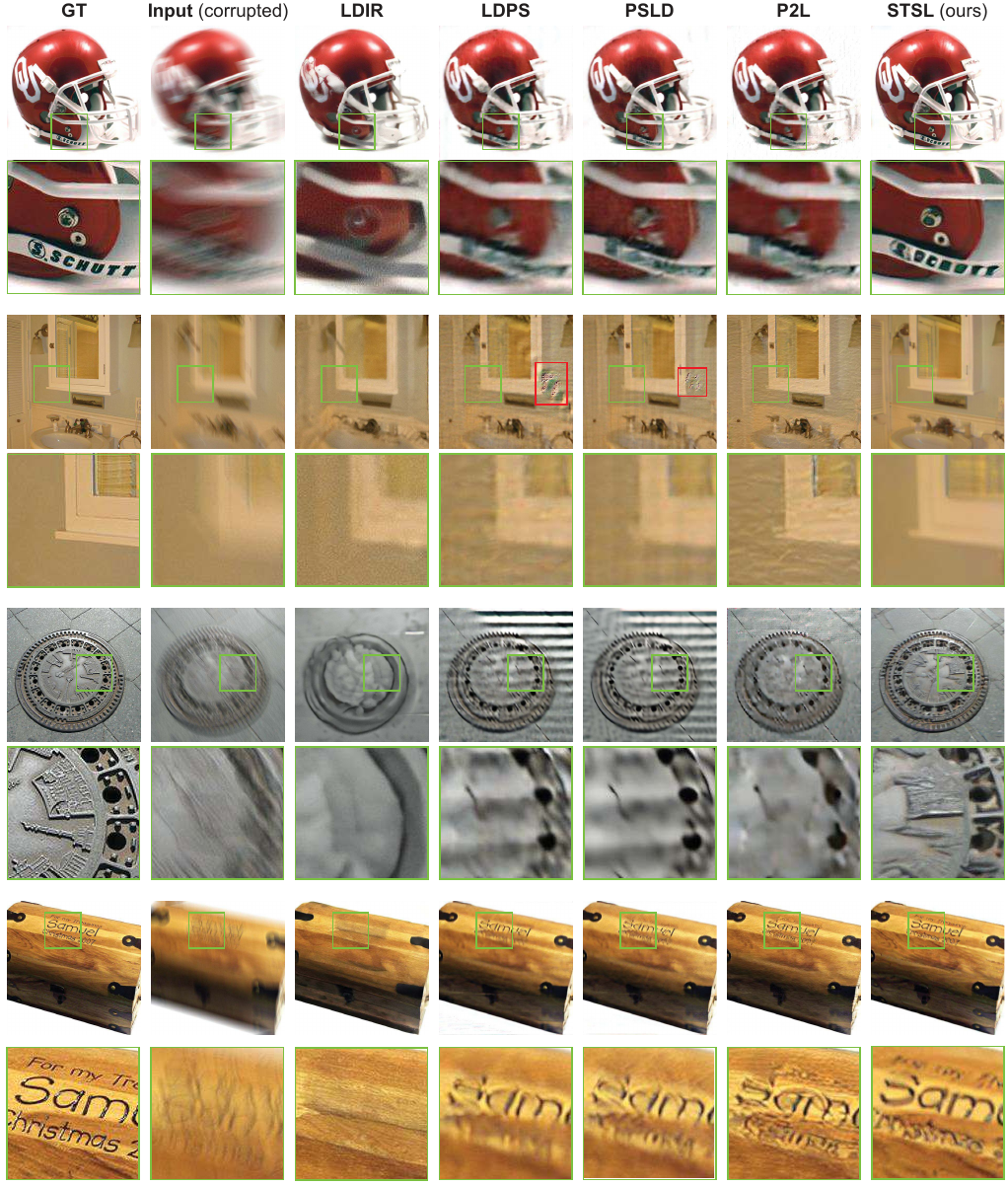}
    \vspace{-4ex}
    \caption{
    \textbf{Qualitative results on motion deblurring:}
    Odd rows represent the full image, while even rows show a zoomed-in view of the {\color{green}{green box}}. 
    The {\color{red}{red boxes}} indicate artifacts from various methods.
    STSL demonstrates superior performance in preserving image details while simultaneously minimizing artifacts and fake textures.
    The competitive baselines: PSLD~\citep{psld} and P2L\citep{p2l} introduce artifacts and fake texture that might be mistaken as sharpness of the reconstructed image.
    Observe the high fidelity text restoration by the proposed approach STSL in the last row.
    }
    \label{fig:appendix:mb}
\end{figure*}

\begin{figure*}
    \centering
    \includegraphics[width=\linewidth]{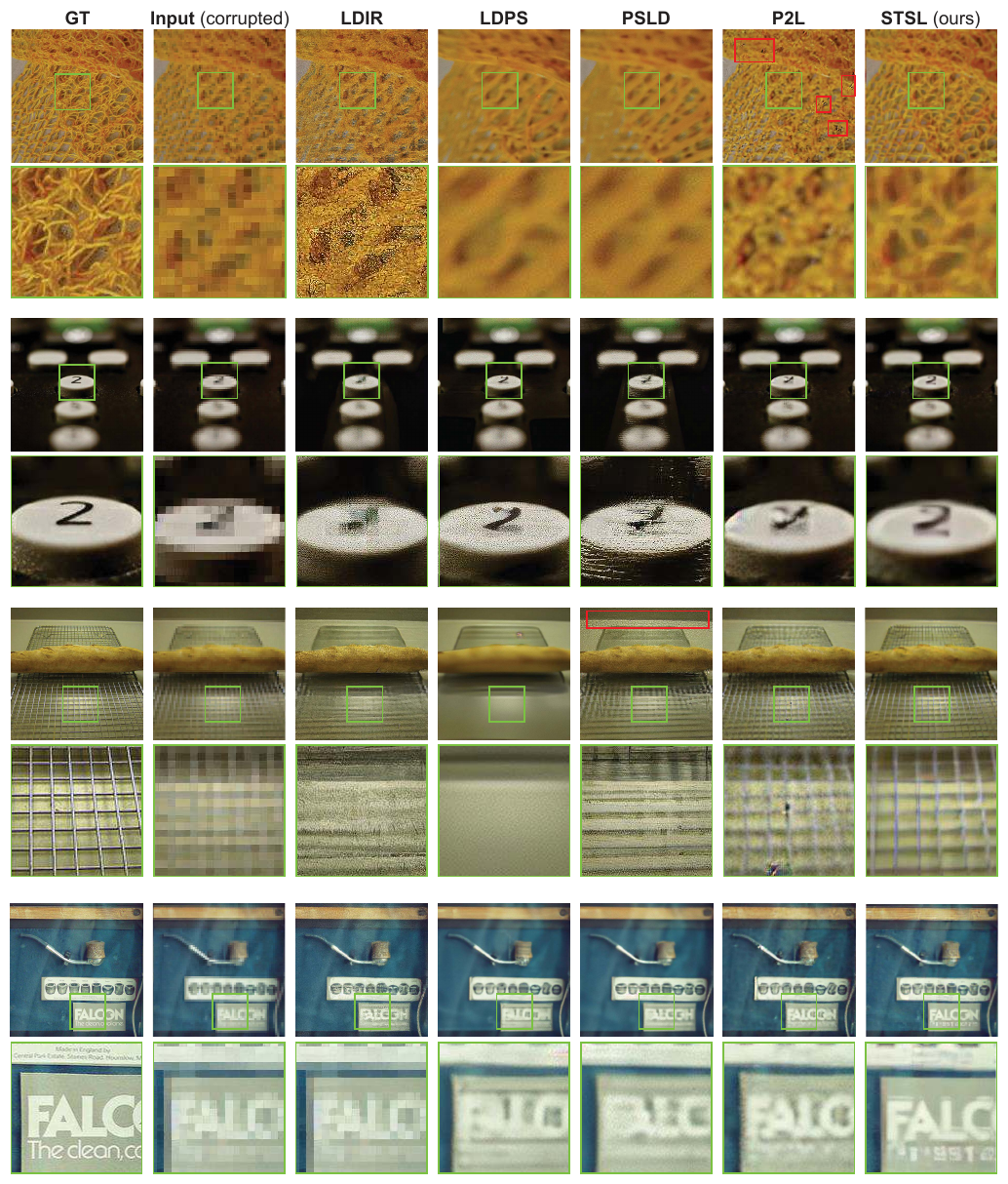}
    \vspace{-4ex}
    \caption{
    \textbf{Qualitative results on SRx8:}
    Odd rows represent the full image, while even rows show a zoomed-in view of the {\color{green}{green box}}. 
    The {\color{red}{red boxes}} indicate artifacts from various methods. STSL restores image details without introducing artifacts (row 1) and shows its potentiality in restoring images with complicated patterns (row 2 and row 6).
    The competitive baselines: PSLD~\citep{psld} and P2L~\citep{p2l} suffer from artifacts that are clearly visible in the highlighted regions. 
    }
    \label{fig:appendix:sr}
\end{figure*}

\begin{figure*}
    \centering
    \includegraphics[width=\linewidth]{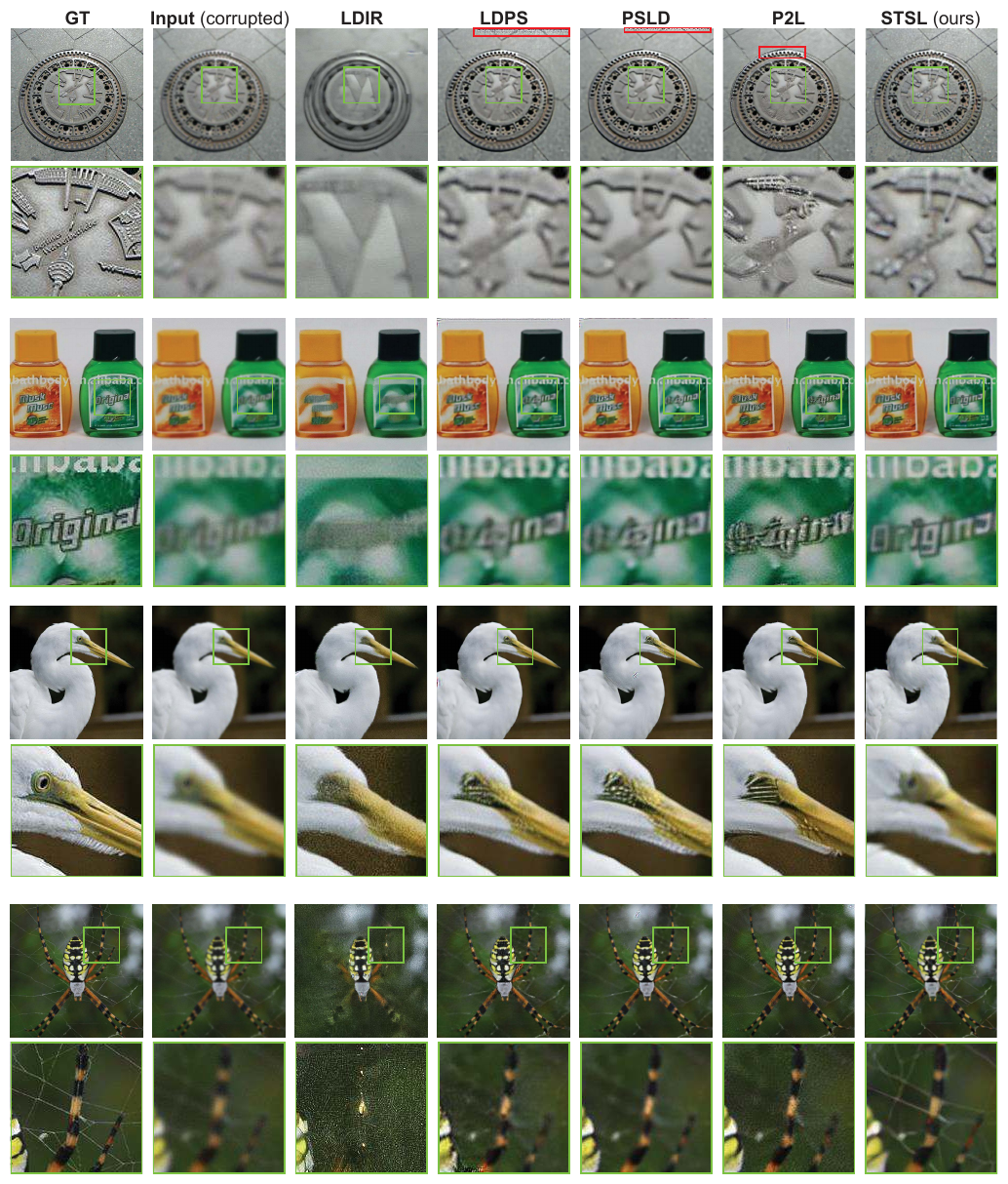}
    \vspace{-4ex}
    \caption{
    \textbf{Qualitative results on Gaussian deblurring:}
    Odd rows represent the full image, while even rows show a zoomed-in view of the {\color{green}{green box}}. 
    The {\color{red}{red boxes}} indicate artifacts from various methods.
    Row 4 and row 8 demonstrate the superior performance of STSL in restoring text and preserving details.
    }
    \label{fig:appendix:gb}
\end{figure*}

\begin{figure*}
    \centering
    \includegraphics[width=\linewidth]{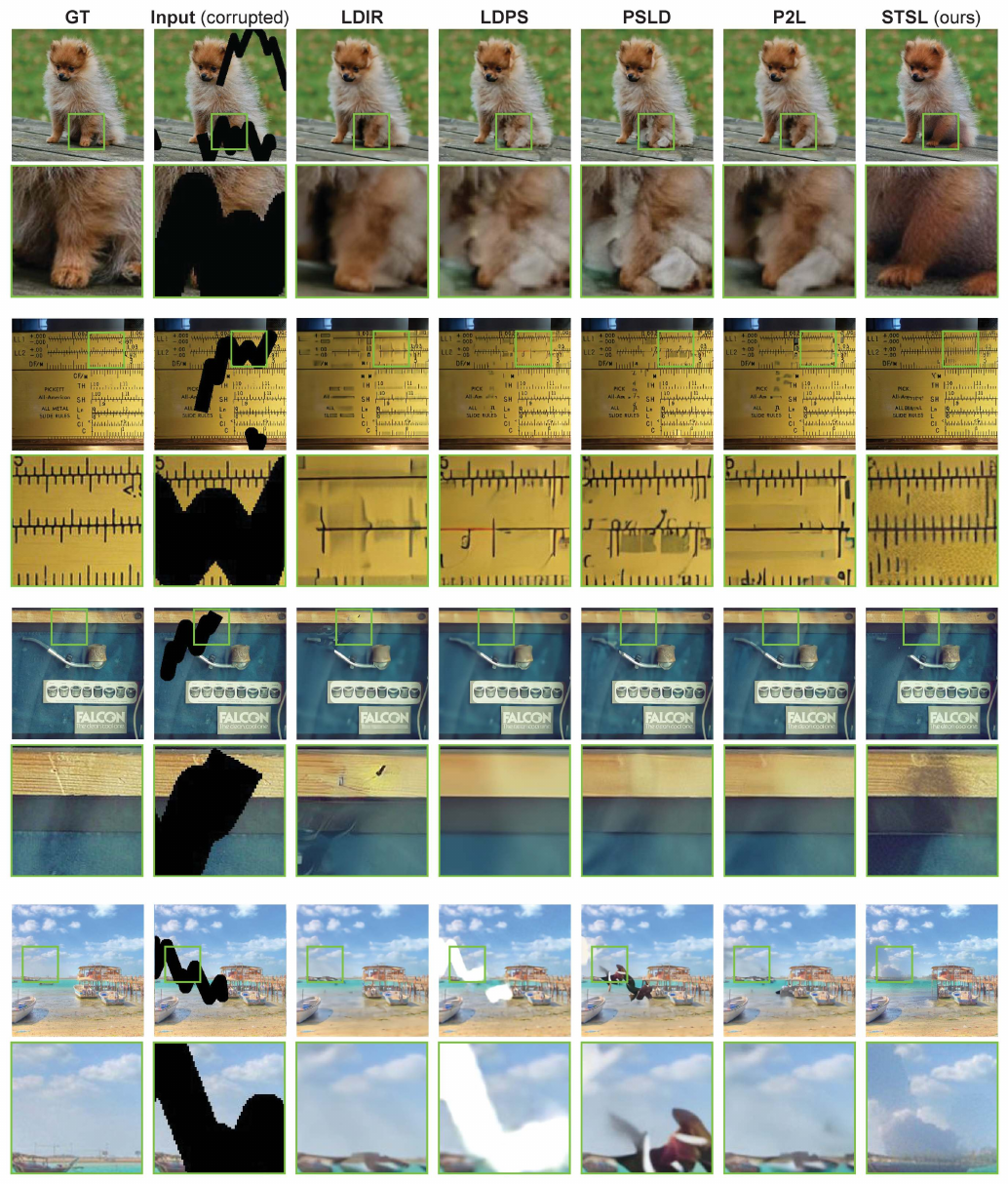}
    \vspace{-4ex}
    \caption{
    \textbf{Qualitative results on free-form inpainting:} Odd rows represent the full image, while even rows show a zoomed-in view of the {\color{green}{green box}}. 
    Note that the model is expected to generate new content that harmonizes with the rest of the pixels, but not necessarily reproduce the same image.
    This is because the goal is to sample the posterior $p\left(X|\vy \right)$.
    The outputs from STSL contain more detailed patterns (row 6) and clear edges (row 2\&4).
    }
    \label{fig:appendix:fip}
\end{figure*}

\begin{figure*}[!t]
    \centering
    \includegraphics[width=\linewidth]{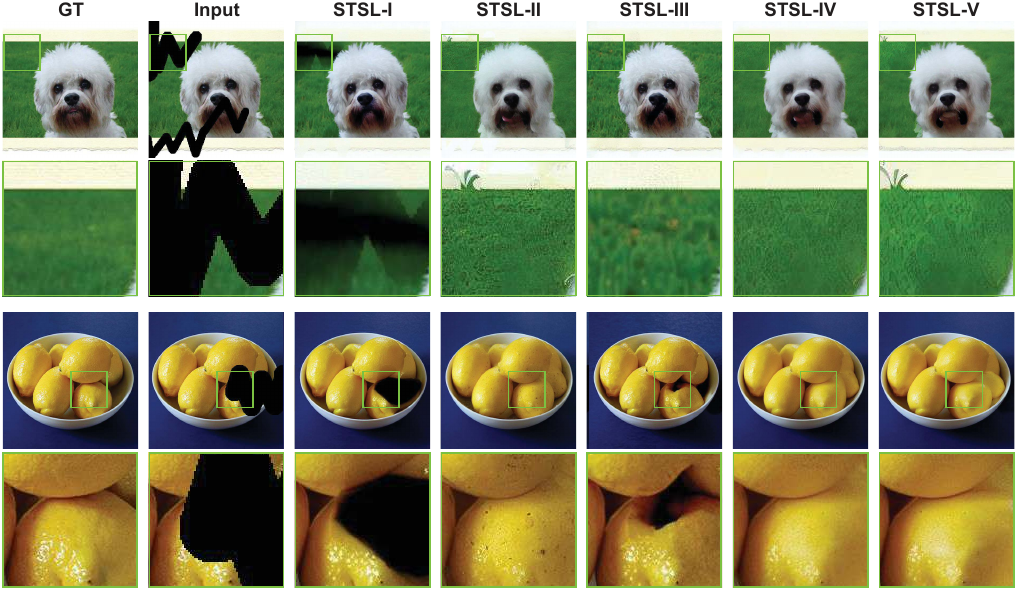}
    \vspace{-4ex}
    \caption{
    \textbf{Failure cases of free-form inpainting:}
    The restored images appear sharp when initialized with the forward latents $Z_0\sim p_T(Z_0|\enc(\mA^T\vy))$ in STSL-I/III, while the images with the reverse process initialized at $Z_0\sim \pi_d$ yield more complete inpainting results (STSL-II/IV/V). One may choose the initialization and the corresponding hyper-parameters as per the requirement in practice. 
    }
    \label{fig:appendix:fip-fail}
\end{figure*}

\begin{figure*}
    \centering
    \includegraphics[width=0.93\linewidth]{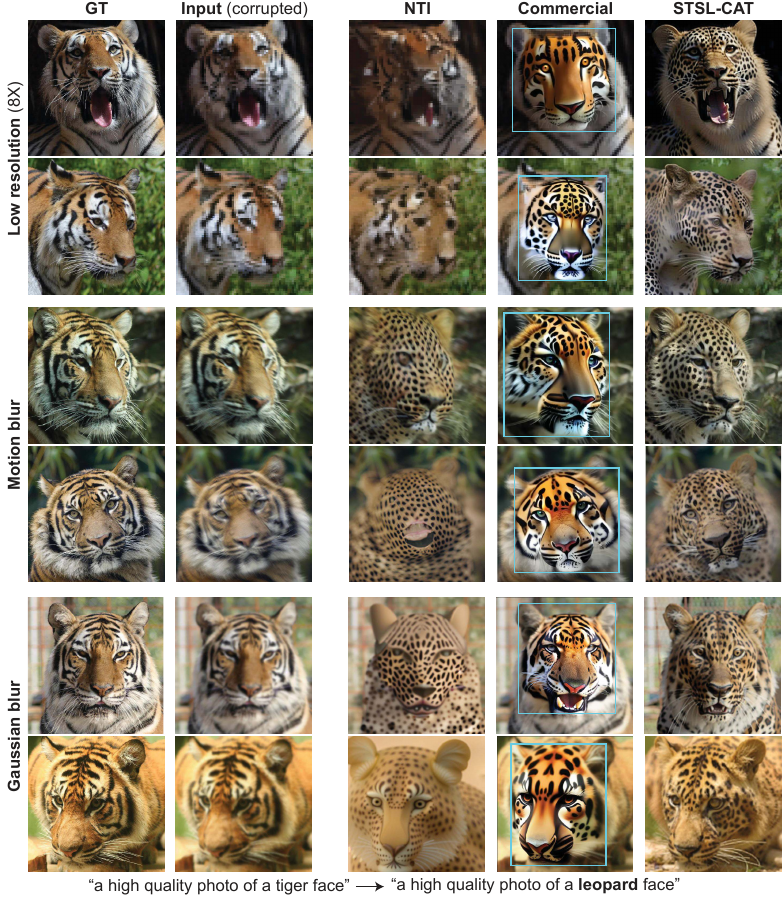}
    \vspace{-1ex}
    \caption{
    \textbf{Qualitative results on image editing on the corrupted images ``tiger'' to ``leopard''.}
    While NTI\citep{nti} fails to conduct high-fidelity image editing when various corruptions are presented, the commercial software synthesizes artistic visual objects without preserving the content of the source image.
    Furthermore, the proposed method STSL-CAT localizes the intended edits without manual intervention, which is necessary for the commercial software. 
    }
    \label{fig:appendix:edit-tiger}
\end{figure*}

\begin{figure*}
    \centering
    \includegraphics[width=0.93\linewidth]{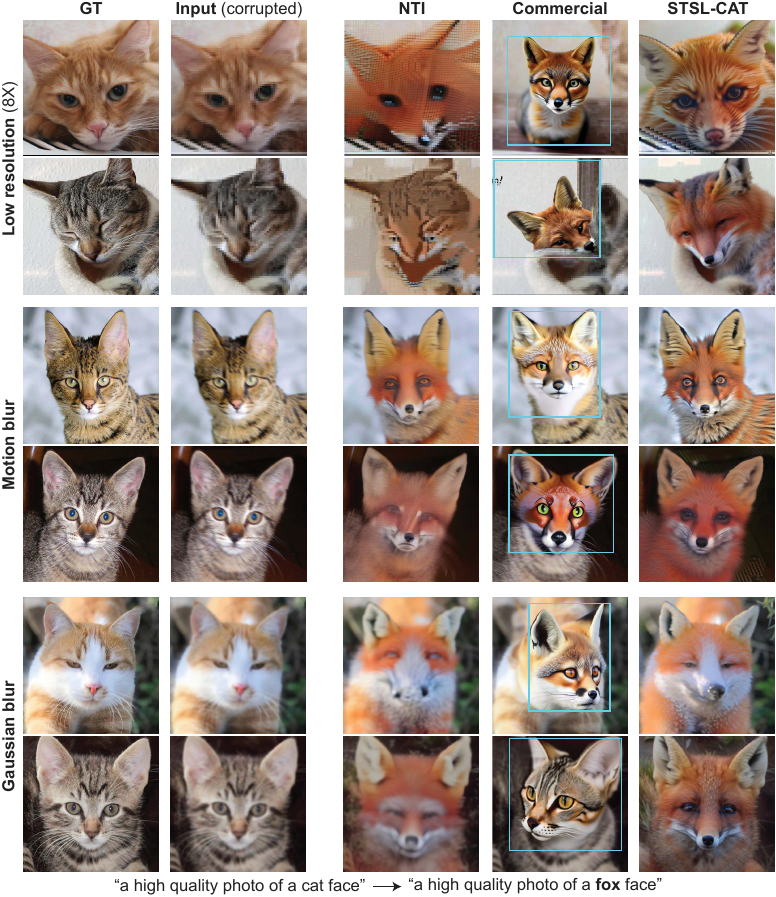}
    \vspace{-1ex}
    \caption{
    \textbf{Qualitative results on image editing on the corrupted images ``cat'' to ``fox''.
    }
    The proposed method STSL preserves the \textit{content} of the source image while performing \textit{text-guided image editing} on corrupt images. 
    }
    \label{fig:appendix:edit-cat}
\end{figure*}

\begin{figure*}
    \centering
    \includegraphics[width=\linewidth]{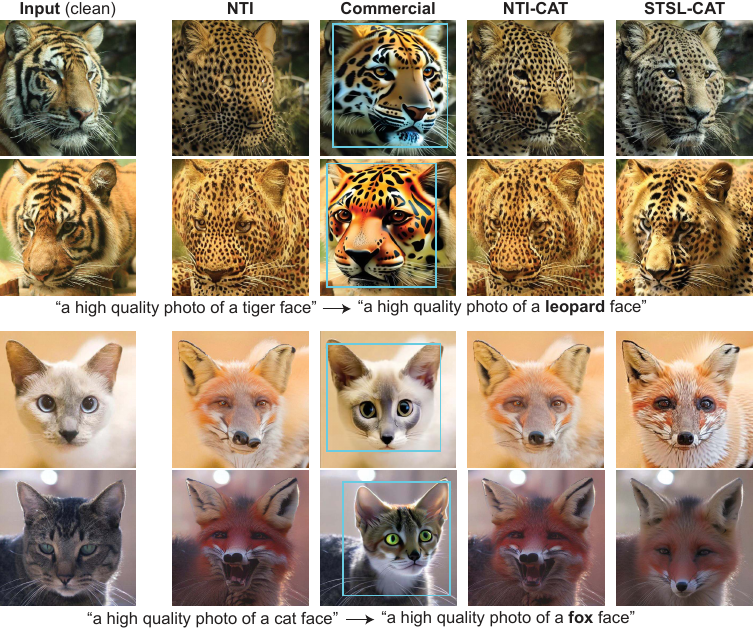}
    \vspace{-1ex}
    \caption{
    \textbf{Qualitative results on Image editing on the clean images.}
    Cross attention tuning (CAT) helps preserve image details with NTI~\citep{nti} (NTI-CAT), and STSL-CAT further enhances the quality of the image by refining the forward latents.
    }
    \label{fig:appendix:edit-clean}
\end{figure*}

\subsection{More Qualitative Results}
\label{sec:more-results}
We present extended results of the proposed method and compare with SoTA solvers in motion deblurring (Figure~\ref{fig:appendix:mb}), SRx8 (Figure~\ref{fig:appendix:sr}), and Gaussian deblurring (Figure~\ref{fig:appendix:gb}).
Notably, STSL demonstrates superior capability in preserving intricate image details and reducing artifact generation, particularly in text-rich images.
This is exemplified in the last images of Figures~\ref{fig:appendix:mb} and \ref{fig:appendix:sr}, where text clarity and legibility are visibly enhanced.
Furthermore, unlike other methods that tend to introduce spurious textures, our approach consistently maintains high image fidelity, reinforcing the effectiveness of STSL in complex scenarios.

Our results also showcase the adaptability of STSL in image editing tasks. In Figures \ref{fig:appendix:edit-tiger} and \ref{fig:appendix:edit-cat}, we illustrate that conventional editing methods struggle with corrupted input images, whereas STSL-CAT achieves high-fidelity editing under these conditions.
Furthermore, STSL-CAT excels in maintaining the integrity of the image even when the input is not corrupted, as demonstrated in Figure~\ref{fig:appendix:edit-clean}.
These qualitative results, supporting the quantitative data presented in Table~\ref{tab:ablation}(f), reveal that the integration of CAT with NTI preserve image \textit{content}, such as in areas of the nose and eyes while changing the \textit{style}. The \textit{refinement} of forward latents (\S\ref{sec:algo-inv}) further contributes to this improvement in rendering details from the corrupt images.

\subsection{Free-form Inpainting}
\label{sec:more-quant-results}
The main body of the paper contains quantitative results on standard datasets. In this section, we provide additional quantitative results on free-form inpainting~\citep{p2l}, which targets to \textit{generate} missing pixels in the blank areas as opposed to \textit{restore} corrupted pixels. Following prior works~\citep{psld,p2l}, we initialize the reverse process at $Z_0\sim \pi_d$ in STSL-II/IV/V. STSL-I/III are initialized at the forward latent $Z_0\sim p_T(Z_0|\enc(\mA^T\vy))$. Table~\ref{tab:appendix} shows the quantitative evaluation on ImageNet ($512\times 512$).
We conduct ablation studies to analyze the latency-optimization trade-offs. As shown in Table~\ref{tab:appendix}, 
STSL uses different combinations of stochastic averaging steps $K$ and DDIM steps $T$.
When compared with methods with the same number of NFEs, STSL outperforms the SoTA solvers PSLD~\citep{psld} and P2L~\citep{p2l} in terms of LPIPS and achieves comparable results in terms of PSNR/SSIM. Figure~\ref{fig:appendix:fip} illustrates the qualitative results on ImageNet.

\noindent\textbf{Limitation:} Figure~\ref{fig:appendix:fip-fail} shows the failure cases of our proposed inverse problem solver STSL in free-form inpainting.
We observe that
the large blocks of missing pixels are embedded into the forward latents 
in STSL-I/III
, which is hard to refine using proximal gradient updates.
Therefore, the masked regions of the final reconstruction sometimes contain incomplete pixels.
This issue arises due to imperfect encoder-decoder of the Stable Diffusion foundation model~\citep{psld},
and could be partly circumvented by slowing down the diffusion process to $T=1000$ steps and initializing the reverse process at $Z_0\sim \pi_d$ as in PSLD~\citep{psld} and P2L~\citep{p2l}. We recommend following this recipe for free-form inpainting.

\end{document}